\documentclass{elsart_new}
\usepackage{graphicx}
\usepackage{amssymb}
\usepackage{amsmath}
\usepackage{color}

\newtheorem{lemma}{Lemma}
\newtheorem{proposition}[lemma]{Proposition}

\newtheorem{theorem}{Theorem}

\newtheorem{result}{Result}

\newenvironment{proof}[1]{\par\noindent{\em Proof#1.}}{\hfill$\Box$\\[2mm]}

\newcommand{\R}{\mathbb{R}}
\newcommand{\N}{\mathbb{N}}

\def\F{\mathcal{F}}

\newcommand{\eps}{\ensuremath{\varepsilon}}
\newcommand{\norm}[1]{\|#1\|}
\DeclareMathOperator{\Bin}{Bin}

\newcommand{\erdos}{Erd{\H{o}}s}
\newcommand{\renyi}{{R{\'e}nyi }}

\newcommand{\Gsym}{G_{\text{sym}}}%
\newcommand{\Gmut}{G_{\text{mut}}}%
\newcommand{\Geps}{G_{\text{eps}}}%
\newcommand{\mut}{_{\text{mut}}}
\newcommand{\sym}{_{\text{sym}}}
\DeclareMathOperator{\kNN}{kNN}
\newcommand{\condon}{\; \boldsymbol{|} \;} %

\newcommand{\supi}{^{(i)}}
\newcommand{\supj}{^{(j)}}
\newcommand{\subi}{_{(i)}}
\newcommand{\minprobu}{\rho}
\newcommand{\maxminprobu}{\rho_{\min}}

\newcommand{\WithinConnEvent}{\mathcal{A}}
\newcommand{\DensDevEvent}{\mathcal{D}}
\newcommand{\ClusterSizeEvent}{\mathcal{B}}
\newcommand{\BoundarySizeEvent}{\mathcal{E}}
\newcommand{\ConnectivityEvent}{\mathcal{C}}
\newcommand{\IsolationEvent}{\mathcal{I}}

\newcommand{\collarSet}{Col}

\newcommand{\kLowerBound}{4^{d+1} \frac{p\supi_{\max}}{t}
  \log \big(2\, 8^d \,p\supi_{\max}\,\vol(C\supi) \, n \big) }
\newcommand{\kUpperBoundN}{(n-1)}
\newcommand{\kUpperBoundWithinConnConst}{2\,4^{d} \, \eta_d \, p\supi_{\max}
  \min\big\{(u\supi)^d,(\nu\supi_{\max})^d\big\}}
\newcommand{\kUpperBoundBetweenConnConst}{\,\frac{\minprobu\supi}{2} -
  \frac{2\log(\tilde{\beta}\subi n)}{\kUpperBoundN} }

\DeclareMathOperator{\vol}{vol}
\DeclareMathOperator{\dist}{dist}

\def\Exp{\mathbb{E}\mathop{\!}\nolimits}
\def\Pr{\mathrm{P}}
\newcommand{\Cluster}{\textrm{Cluster}}
\newcommand{\NoCluster}{\textrm{NoCluster}}

\def\ba#1\ea{\begin{align*}#1\end{align*}}
\def\banum#1\eanum{\begin{align}#1\end{align}}

\usepackage[comma,numbers]{natbib}
\journal{Theoretical Computer Science}
\volume{410}
\pubyear{2009}

\begin{document}

\begin{frontmatter}

\title{Optimal construction of $k$-nearest neighbor graphs
  for identifying noisy clusters\thanksref{tcsarticle}}

\thanks[tcsarticle]{Preprint of an article published in Theoretical 
  Computer Science, Volume 410, Issue 19, Pages 1749--1764, Elsevier, April 2009.}

\author[Tuebingen]{Markus~Maier},
\author[Saarbruecken]{Matthias~Hein},
\author[Tuebingen]{Ulrike~von~Luxburg}
\address[Tuebingen]{Max Planck Institute for Biological Cybernetics, Spemannstr. 38, 72076~T\"{u}bingen, Germany\\ mmaier@tuebingen.mpg.de \hspace{1cm} ulrike.luxburg@tuebingen.mpg.de}
\address[Saarbruecken]{Saarland University, P.O.~Box~151150, 66041~Saarbr\"{u}cken, Germany \\ hein@cs.uni-sb.de}

\begin{abstract}
  We study clustering algorithms based on neighborhood graphs on a
  random sample of data points. The question we ask is how such a
  graph should be constructed in order to obtain optimal clustering
  results. Which type of neighborhood graph should one choose, mutual
  $k$-nearest neighbor or symmetric $k$-nearest neighbor? What is the
  optimal parameter $k$? In our setting, clusters are defined as
  connected components of the $t$-level set of the underlying
  probability distribution. Clusters are said to be identified in the
  neighborhood graph if connected components in the graph correspond
  to the true underlying clusters. Using techniques from random
  geometric graph theory, we prove bounds on the probability that clusters
  are identified successfully, both in a noise-free and in a noisy
  setting. Those bounds lead to several conclusions. First, $k$ has to
  be chosen surprisingly high (rather of the order $n$ than of the
  order $\log n$) to maximize the probability of cluster
  identification. Secondly, the major difference between the mutual
  and the symmetric $k$-nearest neighbor graph occurs when one
  attempts to detect the most significant cluster only.
\end{abstract}

\end{frontmatter}

\section{Introduction} \label{sec-intro}

Using graphs to model real world problems is one of the most
widely used techniques in computer science. This approach usually
involves two major steps: constructing an appropriate graph which
represents the problem in a convenient way, and then constructing
an algorithm which solves the problem on the given type of graph.
While in some cases there exists an obvious natural graph
structure to model the problem, in other cases one has much more
choice when constructing the graph. In the latter cases it is an
important question how the actual construction of the graph
influences the overall result of the graph algorithm.

The kind of graphs we want to study in the current paper are
neighborhood graphs. The vertices of those graphs represent
certain ``objects'', and vertices are connected if the
corresponding objects are ``close'' or ``similar''. The
best-known families of neighborhood graphs are $\eps$-neighborhood
graphs and $k$-nearest neighbor graphs. Given a number of objects
and their mutual distances to each other, in the first case each
object will be connected to all other objects which have distance
smaller than $\eps$, whereas in the second case, each object will be
connected to its $k$ nearest neighbors (exact definitions see below).
Neighborhood graphs are used for modeling purposes in many areas
of computer science: sensor networks and wireless ad-hoc networks,
machine learning, data mining, percolation theory, clustering,
computational geometry, modeling the spread of diseases, modeling
connections in the brain, etc.

In all those applications one has some freedom in constructing
the neighborhood graph, and a fundamental question arises: how
exactly should we construct the neighborhood graph in order to
obtain the best overall result in the end? Which type of
neighborhood graph should we choose? How should we choose its
connectivity parameter, for example the parameter $k$ in the
$k$-nearest neighbor graph? It is obvious that those choices will
influence the results we obtain on the neighborhood graph, but
often it is completely unclear how.

In this paper, we want to focus on the problem of clustering. We
assume that we are given a finite set of data points and pairwise
distances or similarities between them. It is very common to model
the data points and their distances by a neighborhood graph. Then
clustering can be reduced to standard graph algorithms. In the
easiest case, one can simply define clusters as connected
components of the graph. Alternatively, one can try to construct
minimal graph cuts which separate the clusters from each other. An
assumption often made in clustering is that the given data points
are a finite sample from some larger underlying space. For
example, when a company wants to cluster customers based on their
shopping profiles, it is clear that the customers in the company's
data base are just a sample of a much larger set of possible
customers. The customers in the data base are then considered to
be a random sample.

In this article, we want to make a first step towards such results
in a simple setting we call ``cluster identification'' (see next
section for details). Clusters will be represented by connected
components of the level set of the underlying probability
density. Given a finite sample from this density, we want to
construct a neighborhood graph such that we maximize the
probability of cluster identification. To this end, we study
different kinds of $k$-nearest neighbor graphs (mutual, symmetric)
with different choices of $k$ and prove bounds on the
probability that the correct clusters can be identified in this
graph.
One of the first results on the consistency of a clustering method 
has been derived by \citet{Hartigan81}, who proved ``fractional consistency'' 
for single linkage clustering.

The question we want to tackle in this paper is how to choose the
neighborhood graph in order to obtain optimal clustering results.
The mathematical model for building neighborhood graphs on
randomly sampled points is a geometric random graph, see
\citet{Penrose03} for an overview. Such graphs are built by
drawing a set of sample points from a probability measure on $\R^d$,
and then connecting neighboring points (see below for exact
definitions). Note that the random geometric graph model is
different from the classical \erdos-\renyi random graph model (cf.
\citet{Bollobas01} for an overview) where vertices do not have a
geometric meaning, and edges are chosen independently of the
vertices and independently of each other.
In the setup outlined above, the choice of parameter is closely related to
the question of connectivity of random geometric graphs, which
has been extensively studied in the random geometric graph community.
Connectivity results are
not only important for clustering, but also in many other fields
of computer science such as modeling ad-hoc networks (e.g.,
\citet{SanBlo03}, \citet{Bettstetter02}, \citet{KunVen}) or
percolation theory (\citet{BolRio06}). The existing random geometric
graph literature mainly focuses on asymptotic statements about
connectivity, that is results in the limit for infinitely many
data points. Moreover, it is usually assumed that the underlying
density is uniform -- the exact opposite of the setting we consider in
clustering. What we would need in our context are {\em
non-asymptotic} results on the performance of different kinds of
graphs on a {\em finite point set} which has been drawn from
highly clustered densities.

Our results on the choice of graph type and the parameter $k$
for cluster identification can be summarized as follows.
Concerning the question of the choice of $k$, we obtain the
surprising result that $k$ should be chosen surprisingly high,
namely in the order of $O(n)$ instead of
$O(\log n)$ (the latter would be the rate one would ``guess'' from
results in standard random geometric graphs).
Concerning the types of graph, it turns out that different graphs
have advantages in different situations: if one is only interested
in identifying the ``most significant'' cluster (while some
clusters might still not be correctly identified), then the mutual
$\kNN$ graph should be chosen. If one wants to identify many
clusters simultaneously the bounds show no substantial difference
between the mutual and the symmetric $\kNN$ graph.

\section{Main constructions and results}

In this section we give a brief overview over the setup
and techniques we use in the following. Mathematically exact
statements follow in the next sections.

{\bf Neighborhood graphs. } We always assume that we are given $n$
data points $X_1, ..., X_n$ which have been drawn i.i.d. from some
probability measure
which has a density with respect to the Lebesgue measure in $\R^d$.
As distance function between points we use the Euclidean distance,
which is denoted by $\dist$.
The distance is extended to sets $A,B \subseteq \R^d$ via
$\dist(A,B) = \inf \{ \dist(x,y) \condon x \in A, y \in B \}$.
The data points are used as
vertices in an unweighted and undirected graph.
By $\kNN(X_j)$ we denote the set of the $k$
nearest neighbors of $X_j$ among $X_1, ..., X_{j-1}, X_{j+1}, ...,
X_n$. The different neighborhood graphs are defined as follows:
\begin{itemize}
\item
  {\em $\eps$-neighborhood graph $\Geps(n,\eps)$:} $X_i$ and $X_j$ connected
  if $\dist (X_i,X_j)\leq\eps$,
\item
  {\em symmetric $k$-nearest-neighbor graph $\Gsym(n,k)$:}\\
  $X_i$ and $X_j$ connected if $X_i \in \kNN(X_j)$ or $X_j \in \kNN(X_i)$,
\item
  {\em mutual $k$-nearest-neighbor graph $\Gmut(n,k)$:}\\
  $X_i$ and $X_j$ connected if $X_i \in \kNN(X_j)$ and $X_j \in \kNN(X_i)$.
\end{itemize}

Note that the literature does not agree on the names for the
different $\kNN$ graphs. In particular, the graph we call
``symmetric'' usually does not have a special name.

Most questions we will study in the following are much easier to
solve for $\eps$-neighborhood graphs than for
$\kNN$ graphs. The reason is that whether two points $X_i$ and
$X_j$ are connected in the $\eps$-graph only depends on
$\dist (X_i,X_j)$, while in the $\kNN$ graph the existence of an edge
between $X_i$ and $X_j$ also depends on the distances of $X_i$ and
$X_j$ to all other data points. However, the $\kNN$ graph is the
one which is mostly used in practice. Hence we decided to focus on
$\kNN$ graphs. Most of the proofs can easily be adapted
for the $\eps$-graph.

{\bf The cluster model. } There exists an overwhelming amount of
different definitions of what clustering is, and the clustering
community is far from converging on one point of view. In a sample
based setting most definitions agree on the fact that clusters
should represent high density regions of the data space which are
separated by low density regions. Then a straight forward way to define
clusters is to use level sets of the density. Given the
underlying density $p$ of the data space and a parameter $t >0$,
we define the $t$-level set $L(t)$ as the closure of the set of all points $x \in \R^d$
with $p(x) \geq t$. Clusters are then defined as the
connected components of the $t$-level set (where the
term ``connected component'' is used in its topological sense
and not in its graph-theoretic sense).

Note that a different popular model is to define a clustering as a
partition of the whole underlying space such that the boundaries
of the partition lie in a low density area. In comparison, looking
for connected components of $t$-level sets is a stronger
requirement. Even when we are given a complete partition of the
underlying space, we do not yet know which part of each of the
clusters is just ``background noise'' and which one really
corresponds to ``interesting data''. This problem is circumvented by the
$t$-level set definition, which not only distinguishes between the
different clusters but also separates ``foreground'' from
``background noise''. Moreover, the level set approach is much
less sensitive to outliers, which often heavily influence the
results of partitioning approaches.

{\bf The cluster identification problem. }
 Given a finite sample from the underlying distribution, our goal is to
identify the sets of points which come from different connected
components of the $t$-level set. We study this problem in two
different settings:

{\em The noise-free case. } Here we assume that the support of the
density consists of several connected components which have a
positive distance to each other. Between those components, there
is only ``empty space'' (density 0). Each of the connected
components is called a cluster. Given a finite sample $X_1, ...,
X_n$ from such a density, we construct a neighborhood graph $G$
based on this sample. We say that {\bf a cluster is identified in
the graph} if the connected components in the neighborhood graph
correspond to the corresponding connected components of the
underlying density, that is all points originating in the same
underlying cluster are connected in the graph, and they are not
connected to points from any other cluster.

{\em The noisy case. } Here we no longer assume that the clusters are
separated by ``empty space'', but we allow the underlying density to
be supported everywhere. Clusters are defined as the connected
components of the $t$-level set $L(t)$ of the density (for a fixed
parameter $t$ chosen by the user), and points not contained in this level set are
considered as background noise. A point $x \in \R^d$ is called a
{\em cluster point} if $x \in L(t)$ and {\em background point} otherwise.
As in the previous case
we will construct a neighborhood graph $G$ on the given
sample. However, we will remove points from this graph which we
consider as noise. The remaining graph $\tilde G$ will be a subgraph
of the graph $G$, containing fewer vertices and fewer edges than
$G$. As opposed to the noise-free case, we now define two slightly
different cluster identification problems.  They differ in the way
background points are treated. The reason for this more involved
construction is that in the noisy case, one cannot guarantee that no
additional background points from the neighborhood of the cluster will
belong to the graph.

We say that {\bf a cluster is roughly identified} in the remaining
graph $\tilde G$ if the following properties hold: \vspace{-0.4cm}
  \begin{itemize}
  \item
    all sample points from a cluster are contained as vertices
    in the graph, that is, only background points are dropped,
  \item
    the vertices belonging to the same cluster are connected in the
    graph, that is, there exists a path between each two of them, and
  \item
    every connected component of the graph contains only points of
    exactly one cluster (and maybe some additional noise points, but
    no points of a different cluster).
  \end{itemize}
We say that {\bf a cluster is exactly identified in $\tilde G$} if
\vspace{-0.4cm}
\begin{itemize}
\item it is roughly identified, and \item the ratio of the number
of background points and the number of
  cluster points in the graph $\tilde G$ converges almost surely to zero as the
  sample size approaches infinity.
\end{itemize}

If all clusters have been roughly identified, the number of
connected components of the graph $\tilde G$ is equal to the
number of connected components of the level set $L(t)$. However,
the graph $\tilde G$ might still contain a significant number of
background points. In this sense, exact cluster identification
is a much stronger problem, as we require that the fraction of
background points in the graph has to approach zero.
Exact cluster identification is an asymptotic statement,
whereas rough cluster identification can be verified on each
finite sample. Finally, note that in the noise-free case,
rough and exact cluster identification coincide.

{\bf The clustering algorithms. } To determine the clusters in the
finite sample, we proceed as follows. First, we construct a
neighborhood graph on the sample. This graph looks different,
depending on whether we allow noise or not:

{\em Noise-free case. } Given the data, we simply construct the
mutual or symmetric $k$-nearest neighbor graph
($G\mut (n,k)$ resp. $G\sym (n,k)$) on the data points,
for a certain parameter $k$, based on the Euclidean distance.
Clusters are then the connected components of this graph.

{\em Noisy case. } Here we use a more complex procedure:
\begin{itemize}
\item
  As in the noise-free case, construct the mutual (symmetric)
  $\kNN$ graph $G\mut (n,k)$ (resp. $G\sym (n,k)$) on the samples.
\item
  Estimate the density
  $\hat p_n ( X_i )$ at every sample point $X_i$ (e.g., by kernel density estimation).
\item
  If $\hat p_n ( X_i ) < t'$, remove the point
  $X_i$ and its adjacent edges from the graph (where $t'$ is a
  parameter determined later).
  The resulting graph is denoted by
  $G\mut' \left( n, k, t' \right)$ (resp. $G\sym' \left( n, k, t' \right)$).
\item
  Determine the connected components of
  $G\mut' \left( n, k, t' \right)$ (resp. $G\sym' \left( n, k, t' \right)$),
  for example by a simple depth-first search.
\item
  Remove the connected components of the graph that are ``too small'',
  that is, which contain less than $\delta n$ points
  (where $\delta$ is a small parameter determined later).
\item
  The resulting graph is denoted by
  $\tilde{G}\mut \left( n, k, t', \delta \right)$
  (resp. $\tilde{G}\sym \left( n, k, t', \delta \right)$);
  its connected components are the clusters of the sample.
\end{itemize}
Note that by removing the small components in the graph the method
becomes very robust against outliers and ``fake'' clusters (small
connected components just arising by random fluctuations).

{\bf Main results, intuitively. } \label{par:intuitive_results}
We would like to outline our results briefly in an intuitive
way. Exact statements can be found in the following sections.

\begin{result}[Range of $k$ for successful cluster identification]
  Under mild assumptions, and for $n$ large enough, there exist
  constants $c_1,c_2 >0$ such that for any $k \in [c_1 \log n, c_2 n]$,
  all clusters are identified with high probability in both
  the mutual and symmetric $\kNN$ graph. This result holds for cluster
  identification in the noise-free case as well as for the rough and
  the exact cluster identification problem (the latter seen as an asymptotic statement)
  in the noisy case (with different constants $c_1,c_2$).
\end{result}

For the noise-free case, the lower bound on $k$ has already been
proven in \citet{BriChaQuiYuk1997}, for the noisy case it is new.
Importantly, in the exact statement of the result all constants
have been worked out more carefully than in
\citet{BriChaQuiYuk1997}, which is very important for proving the
following statements.

\begin{result} {\bf (Optimal $k$ for cluster identification)}
Under mild assumptions, and for $n$ large enough, the parameter
$k$ which maximizes the probability of successful identification
of one cluster in the noise-free case has the form $k = c_1 n + c_2$,
where $c_1, c_2$ are constants which
depend on the geometry of the cluster. This result holds for both
the mutual and the symmetric $\kNN$ graph, but the convergence
rates are different (see Result~3). A similar result holds
as well for rough cluster identification in the noisy case, with
different constants.
\end{result}

This result is completely new, both in the noise-free and in the
noisy case. In the light of the existing literature, it is rather
surprising. So far it has been well known that in many different
settings the lower bound for obtaining connected components in a
random $\kNN$ graph is of the order $k \sim \log n$. However, we
now can see that {\em maximizing the probability} of obtaining connected
components on a finite sample leads to a dramatic change: $k$ has
to be chosen much higher than $\log n$, namely of the order
$n$ itself. Moreover, we were surprised ourselves that this result
does not only hold in the noise-free case, but can also be carried
over to rough cluster identification in the noisy setting.

For exact cluster identification we did not manage to determine an
optimal choice of $k$ due to the very difficult setting. For large
values of $k$, small components which can be discarded will no longer
exist.  This implies that a lot of background points are attached to
the real clusters. On the other hand, for small values of $k$ there
will exist several small components around the cluster which are
discarded, so that there are less background points attached to the
final cluster. However, this tradeoff is very hard to grasp in
technical terms. We therefore leave the determination of an optimal
value of $k$ for exact cluster identification as an open problem.
Moreover, as exact cluster identification concerns the
asymptotic case of $n \to \infty$ only, and rough cluster
identification is all one can achieve on a finite sample anyway,
we are perfectly happy to be able to prove the optimal rate in that case.

\begin{result}[Identification of the most significant cluster]
For the optimal $k$ as stated in Result~2, the convergence rate
(with respect to $n$)
for the identification of one fixed cluster $C\supi$ is
different for the mutual and the symmetric $\kNN$ graph.
It depends \vspace{-0.4cm}
\begin{itemize}
\item
  only on the properties of the cluster $C\supi$ itself in the mutual
  $\kNN$ graph \item on the properties of the ``least significant'',
  that is the ``worst'' out of all clusters in the symmetric
  $\kNN$ graph.
\end{itemize}
\end{result}

This result shows that if one is interested in identifying the
``most significant'' clusters only, one is better off
using the mutual $\kNN$ graph. When the goal is to identify all
clusters, then there is not much difference between the two graphs,
because both of them have to deal with the ``worst'' cluster
anyway. Note that this result is mainly due to the different
between-cluster connectivity properties of the graphs, the
within-cluster connectivity results are not so different
(using our proof techniques at least).

{\bf Proof techniques, intuitively. } Given a neighborhood graph
on the sample, cluster identification always consists of two main
steps: ensuring that points of the same cluster are connected and
that points of different clusters are not connected to each other.
We call those two events ``within-cluster connectedness'' and
``between-cluster disconnectedness'' (or ``cluster isolation'').

To treat within-cluster connectedness we work with a covering
of the true cluster. We cover
the whole cluster by balls of a certain radius $z$. Then we want
to ensure that, first, each of the balls contains at least one of
the sample points, and second, that points in neighboring balls
are always connected in the $\kNN$ graph. Those are two
contradicting goals. The larger $z$ is, the easier it is to
ensure that each ball contains a sample point. The smaller $z$ is,
the easier it is to ensure that points in neighboring balls will
be connected in the graph for a fixed number of neighbors $k$.
So the first part of the proof consists
in computing the probability that for a given $z$ both events
occur at the same time and finding the optimal $z$.

Between-cluster connectivity is easier to treat.
Given a lower bound on the distance $u$ between two clusters, all
we have to do is to make sure that edges in the $\kNN$ graph never
become longer than $u$, that is we have to prove bounds on the
maximal $\kNN$ distance in the sample.

In general, those techniques can be applied with small
modifications both in the noise-free and in the noisy case,
provided we construct our graphs in the way described above. The
complication in the noisy case is that if we just used the
standard $\kNN$ graph as in the noise-free case, then of course
the whole space would be considered as one connected component,
and this would also show up in the neighborhood graphs. Thus, one has
to artificially reduce the neighborhood graph in order to remove the
background component. Only then one can hope to obtain a graph
with different connected components corresponding to different
clusters. The way we construct the graph $\tilde G$ ensures this.
First, under the assumption that the error of the density
estimator is bounded by $\eps$, we consider the $(t -
\eps)$-level set instead of the $t$-level set we are interested
in. This ensures that we do not remove ``true cluster points'' in
our procedure. A second, large complication in the noisy case is
that with a naive approach, the radius $z$ of the covering and the
accuracy $\eps$ of the density estimator would be coupled to each
other. We would need to ensure that the parameter $\eps$ decreases
with a certain rate depending on $z$. This would lead to complications
in the proof as well as very slow convergence rates. The trick by which we can avoid this
is to introduce the parameter $\delta$ and throw away all
connected components which are smaller than $\delta n$. Thus, we
ensure that no small connected components are left over in the
boundary of the $(t-\eps)$-level set of a cluster, and all remaining points
which are in this boundary strip will be connected to the main cluster
represented by the $t$-level set. Note, that this construction allows us to
estimate the number of clusters even without exact estimation of
the density.

{\bf Building blocks from the literature. } To a certain extent,
our proofs follow and combine some of the techniques presented in
\citet{BriChaQuiYuk1997} and \citet{BiaCadPel:2007}.

In \citet{BriChaQuiYuk1997} the authors study the connectivity of
random mutual $k$-nearest neighbor graphs. However, they are
mainly interested in asymptotic results, only consider the
noise-free case, and do not attempt to make statements about the
optimal choice of $k$. Their main result is that in the noise-free
case, choosing $k$ at least of the order $O(\log n)$ ensures that
in the limit for $n \to \infty$, connected components of the
mutual $k$-nearest neighbor graph correspond to true underlying
clusters.

In \citet{BiaCadPel:2007}, the authors study the noisy case and
define clusters as connected components of the $t$-level set of
the density. As in our case, the authors use density estimation to
remove background points from the sample, but then work with an
$\eps$-neighborhood graph instead of a $k$-nearest neighbor graph
on the remaining sample. Connectivity of this kind of graph is
much easier to treat than the one of $k$-nearest neighbor graphs,
as the connectivity of two points in the $\eps$-graph does not depend on any other
points in the sample (this is not the case in the $k$-nearest
neighbor graph). Then, \citet{BiaCadPel:2007} prove asymptotic results for the
estimation of the connected components of the level set $L(t)$,
but also do not investigate the optimal choice of their graph
parameter $\eps$. Moreover, due to our additional step where we remove
small components of the graph, we can
provide much faster rates for the estimation of the components, since
we have a much weaker coupling of the density estimator and the clustering
algorithm.

Finally, note that a considerably shorter version of the current
paper dealing with the noise-free case only has appeared
in~\citet{OurALTPaper:2007}. In the current paper we have shortened the proofs
significantly at the expense of having slightly worse constants
in the noise-free case.

\section{General assumptions and notation}

{\bf Density and clusters. } Let $p$ be a bounded probability
density with respect to the Lebesgue measure on $\R^d$. The
measure on $\R^d$ that is induced by the density $p$ is denoted by $\mu$.
Given a fixed level parameter $t>0$, the
$t$-level set of the density $p$ is defined as \ba L(t) =
\overline{\{ x \in \R^d \condon p(x) \geq t\}}. \ea where the bar
denotes the topological closure (note that level sets are
closed by assumptions in the noisy case, but this is not
necessarily the case in the noise-free setting).

{\bf Geometry of the clusters. } We define clusters as
the connected components of $L(t)$ (where the term
``connected component'' is used in its topological sense).
The number of clusters is denoted by $m$, and the clusters
themselves by $C^{(1)}, \ldots,C^{(m)}$.
We set $\beta\subi := \mu ( C^{(i)} )$, that means, the
probability mass in cluster $ C^{(i)}$.

We assume that each cluster $C\supi$ ($i=1, \ldots ,m$) is a
disjoint, compact and connected subset of $\R^d$,
whose boundary $\partial C^{(i)}$ is a smooth $(d-1)$-dimensional
submanifold in $\R^d$ with minimal curvature radius
$\kappa^{(i)}>0$ (the inverse of the largest principal curvature
of $\partial C\supi$).
For $\nu \leq \kappa^{(i)}$, we define the collar set
$\collarSet^{(i)}( \nu ) = \{x \in C^{(i)} \, \big|\, \dist(x,\partial C^{(i)})\leq \nu \}$
and the maximal covering radius
$\nu^{(i)}_{\max} =
\max_{\nu \leq \kappa^{(i)}} \{\nu \; | \;
C^{(i)}\setminus \collarSet^{(i)}(\nu) \; \textrm{connected}\;\}$.
These quantities will be needed for the following
reasons: It will be necessary to cover the inner part of each
cluster by balls of a certain fixed radius $z$, and those balls
are not supposed to ``stick outside''. Such a construction is only
possible under assumptions on the maximal curvature of the
boundary of the cluster. This will be particularly important in
the noisy case, where all statements about the density estimator
only hold in the inner part of the cluster.

\begin{figure}[ht]
  \begin{center}
    \includegraphics{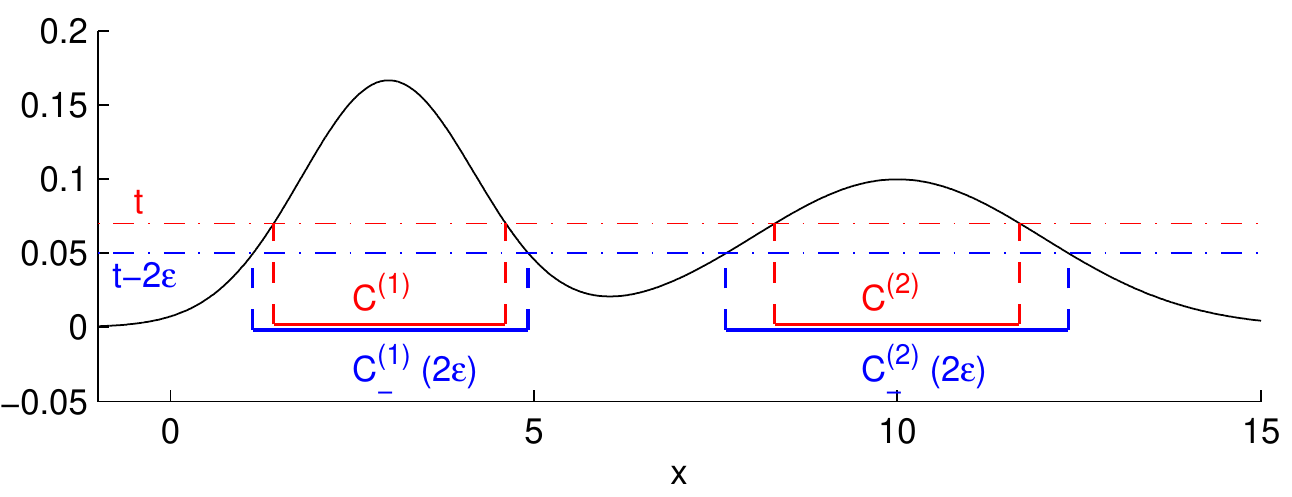}
    \caption{\label{fig:cluster_definition} An example of our cluster
      definition. The clusters $C^{(1)}$, $C^{(2)}$ are defined as the
      connected components of the $t$-level set of the density
      (here $t=0.07$).
      The clusters are subsets of the sets
      $C_-^{(1)} (2\epsilon)$, $C_-^{(2)}(2\epsilon)$
      (here for $\epsilon=0.01$).}
  \end{center}
\end{figure}

For an arbitrary $\eps> 0$, the connected component of $L(t-\eps)$
which contains the cluster $C^{(i)}$ is denoted by $C_-\supi (\eps)$.
Points in the set $ C_-\supi (\eps) \setminus C^{(i)}$ will
sometimes be referred to as boundary points.
To express
distances between the clusters, we assume that there exists some
$\tilde{\eps} > 0$ such that $\dist( C^{(i)}_- (2 \tilde{\eps}
), C^{(j)}_- ( 2 \tilde{\eps} )) \geq u\supi > 0$ for all $i,j
\in \{1, \ldots ,m \}$. The numbers $u\supi$ will represent lower
bounds on the distances between cluster $C\supi$ and the remaining
clusters. Note that the existence of the $u\supi > 0$ ensures that
$C_-\supi (2\eps)$ does not contain any other clusters apart
from $C^{(i)}$ for $\eps<\tilde{\eps}$.
Analogously to the definition of $\beta\subi$ above we set
$\tilde{\beta}\subi = \mu ( C_-^{(i)} ( 2 \tilde{\eps} ))$,
that is the mass of the enlarged set $ C_-^{(i)} ( 2 \tilde{\eps})$.
These definitions are illustrated in Figure~\ref{fig:cluster_definition}.
Furthermore, we introduce a lower bound on the probability mass in
balls of radius $u\supi$ around points in $C\supi_-(2 \tilde{\eps})$
\begin{displaymath}
  \minprobu\supi \leq \inf_{x \in C\supi_-(2 \tilde{\eps})}
  \mu \left( B(x,u\supi) \right) .
\end{displaymath}
In particular, under our assumptions on the smoothness of the cluster boundary
we can set $\minprobu\supi= O\supi(u\supi) t \eta_d( u\supi )^d$
for an {\em overlap constant}
\ba
O^{(i)}(u\supi)=\inf\limits_{x \in C^{(i)}_- ( 2 \tilde{\eps} )} \big(
\vol ( B(x,u\supi) \cap C^{(i)}_- ( 2\tilde{\eps} ) ) / \vol
  ( B(x,u\supi) ) \big) > 0 .
\ea
The way it is constructed, $\minprobu\supi$ becomes
larger the larger the distance of $C\supi$ to all the other clusters
is and is upper bounded by the probability mass of the extended cluster $\tilde{\beta}\subi$.

{\em Example in the noisy case. } All assumptions on the density
and the clusters are satisfied if we assume that the density $p$
is twice continuously
differentiable on a neighborhood of
$\{ p = t \}$, for each $x \in \{ p= t \}$ the gradient of
$p$ at $x$ is non-zero, and
$\dist ( C^{(i)}, C^{(j)} ) = u' > u^{(i)}$.

{\em Example in the noise-free case. } Here we assume that the
support of the density $p$ consists of $m$ connected components $C^{(1)}, \ldots
,C^{(m)}$ which satisfy the smoothness assumptions above, and such
that the densities on the connected components are lower bounded
by a positive constant $t$. Then the noise-free case is a special
case of the noisy case.

{\bf Sampling. } Our $n$ sample points $X_1, ..., X_n$ will be sampled
i.i.d. from the underlying probability distribution.

{\bf Density estimation in the noisy case. } In the noisy case we
will estimate the density at each data point $X_j$ by some estimate
$\hat p_n(X_j)$. For convenience, we state some of our results using a
standard kernel density estimator, see \citet{DevLug01} for background
reading. However, our results can easily be rewritten with any other
density estimate.

{\bf Further notation. } The $\kNN$ radius of a point $X_j$ is the
maximum distance to a point in $\kNN(X_i)$. $R^{(i)}_{\min}$
denotes the minimal $\kNN$ radius of the sample points in cluster
$C^{(i)}$, whereas $\tilde{R}^{(i)}_{\max}$ denotes the maximal $\kNN$
radius of the sample points in $C_-^{(i)} (2 \tilde{\eps})$.
Note here the difference in the point sets that are considered.

$\Bin ( n,p )$ denotes the binomial distribution with parameters
$n$ and $p$.
Probabilistic events will be denoted with curly capital letters
$\mathcal{A}, \mathcal{B}, \hdots$, and their complements with
$\mathcal{A}^c, \mathcal{B}^c, \hdots$.

\begin{table}[h]
  \caption{Table of notations}
  \centering
  \begin{tabular}{|l|l|}
    \hline
    $p(x)$ & density \\
    $\hat{p}_n (x)$ & density estimate in point $x$ \\
    $t$ & density level set parameter \\
    $L(t)$ & $t$-level set of $p$ \\
    $C^{(1)}, \ldots ,C^{(m)}$ & clusters, i.e. connected components of $L(t)$ \\
    $C\supi_- (\epsilon)$ & connected component of $L(t-\epsilon)$
    containing $C\supi$ \\
    $\beta\subi$, $\tilde{\beta}\subi$ & probability mass of $C^{(i)}$ and
    $C^{(i)}_- (2 \tilde{\eps})$ respectively \\
    $p^{(i)}_{\max}$ & maximal density in cluster $C\supi$ \\
    $\minprobu\supi$ & probability of balls of radius $u\supi$ around points in
    $C^{(i)}_- (2 \tilde{\eps})$ \\
    $\kappa \supi$ & minimal curvature radius of the boundary $\partial C\supi$ \\
    $\nu^{(i)}_{\max}$ & maximal covering radius of cluster $C^{(i)}$\\
    $\collarSet^{(i)}( \nu )$ & collar set for radius $\nu$ \\
    $u^{(i)}$ & lower bound on the distances between $C\supi$ and other clusters \\
    $\tilde{\epsilon}$ & parameter such that
    $\dist( C^{(i)}_- (2 \eps), C^{(j)}_- ( 2 \eps )) \geq u\supi$
    for all $\eps \leq \tilde{\eps}$ \\
    $\eta_d$ & volume of the $d$-dimensional unit ball \\
    $k$ & number of neighbors in the construction of the graph \\
    \hline
  \end{tabular}
\end{table}

\section{Exact statements of the main results}

In this section we are going to state all our main results in a formal way.
In the statement of the theorems we need the following
conditions. The first one is necessary for both,
the noise-free and the noisy case, whereas the second one is needed
for the noisy case only.
\begin{itemize}
\item
  {\em Condition 1:} Lower and upper bounds on the number of neighbors $k$,
  \begin{align*}
    k\,&\geq\,  \kLowerBound ,\\
    k\,&\leq\,  \kUpperBoundN \min\Big\{\kUpperBoundBetweenConnConst,
    \kUpperBoundWithinConnConst \Big\}.
  \end{align*}
\item
  {\em Condition 2:}
  The density $p$ is three times continuously differentiable with uniformly bounded 
  derivatives,
  $\beta\subi > 2 \delta$, and
  $\eps_n$ sufficiently small such that
  \sloppy $\mu \big(\bigcup_i ( C_-\supi(2\eps_n)\backslash C\supi ) \big)\leq \delta/2$.
\end{itemize}
Note that in Theorems~1 to~3 $\eps_n$ is considered small but constant and
thus we drop the index $n$ there.

In our first theorem, we present the optimal choice of the parameter $k$ in the mutual $\kNN$ graph
for the identification of a cluster. This theorem treats both,
the noise-free and the noisy case.

\begin{theorem} {\bf (Optimal $k$ for identification of one
    cluster in the mutual $\kNN$ graph)} \label{cor:optimal-k-noise}
  The optimal choice of $k$ for identification of
  cluster $C\supi$ in $\Gmut(n,k)$ (noise-free case) resp.
  rough identification in
  $\tilde{G}\mut \left( n, k, t - \eps, \delta \right)$ (noisy case) is
  \begin{align*}
    k = (n-1)\Gamma\supi +1, \quad  \textrm{ with }  \quad \Gamma\supi:= \frac{\minprobu\supi}{2 + \frac{1}{4^d}\frac{t}{p\supi_{\max}}},
  \end{align*}
  provided this choice of $k$ fulfills Condition~1.\\
  In the noise-free case we obtain with
  $\Omega\supi_{\textrm{noisefree}} =\frac{\minprobu\supi}{2\,4^{d+1}\,\frac{p\supi_{\max}}{t} + 4}$ and for sufficiently large $n$
  \[ \Pr \big(\textrm{Cluster $C\supi$ is identified in $\Gmut(n,k)$} \big) \geq 1 - 3 e^{-(n-1)\Omega\supi_{\textrm{noisefree}}}.\]
  For the noisy case, assume that additionally Condition~2 holds and
  let $\hat{p}_n$ be a kernel density estimator with bandwidth $h$.
  Then there exist constants $C_1,C_2$ such that if $h^2\leq C_1\eps$ we get with
  \[ \Omega\supi_{\textrm{noisy}} = \min\bigg\{\frac{\minprobu\supi}{2\,4^{d+1}\,\frac{p\supi_{\max}}{t} + 4}, \; \frac{n}{n-1}\frac{\delta}{8}, \; \frac{n}{n-1}C_2\, h^d\,\eps^2\bigg\}\]
  and for sufficiently large $n$
  \[ \Pr \big( \textrm{Cluster $C\supi$ roughly identified in $\tilde{G}\mut \left( n, k, t - \eps, \delta \right)$} \big) \geq 1 - 8 e^{-(n-1)\Omega\supi_{\textrm{noisy}}}.\]
\end{theorem}

This theorem has several remarkable features. First of all, we can see
that both in the noise-free and in the noisy case, the optimal choice
of $k$ is roughly linear in $n$. This is pretty surprising, given that
the lower bound for cluster connectivity in random geometric graphs
is $k \sim \log n$. We will discuss the important consequences of this
result in the last section.

Secondly, we can see that for the mutual $\kNN$ graph the
identification  of one cluster $C\supi$ only depends on the
properties of the cluster $C\supi$, but not on the ones of any other
cluster. This is a unique feature of the mutual $\kNN$ graph which
comes from the fact that if cluster $C\supi$ is very ``dense'', then
the neighborhood relationship of points in $C\supi$ never links
outside of cluster $C\supi$.  In the mutual $\kNN$ graph this implies that any
connections of $C\supi$ to other clusters are prevented. Note that
this is not true for the symmetric $\kNN$ graph, where another cluster
can simply link into $C\supi$, no matter which internal properties
$C\supi$ has.

For the mutual graph, it thus makes
sense to  define the {\em most significant} cluster as the one with
the largest coefficient $\Omega\supi$, since this is the one which can be identified with the fastest rate.
In the noise-free case one
observes that the coefficient $\Omega\supi$ of cluster $C\supi$ is
large given that
\vspace{-0.4cm}
\begin{itemize}
\item $\minprobu\supi$ is large, which effectively  means a large distance $u\supi$ of
  $C\supi$ to the closest other cluster,
\item  $p\supi_{\max}/ t$ is small, so that the density is rather uniform inside the
cluster $C\supi$..
\end{itemize}
\vspace{-0.4cm}
Note that those properties are the most simple properties one would
think of when imagining an ``easily detectable'' cluster.
For the noisy case, a similar analysis still holds as long as one can choose
the constants $\delta,h$ and $\eps$ small enough.

Formally, the result for identification of clusters in the symmetric
$\kNN$ graph looks very similar to the one above.
\begin{theorem} {\bf (Optimal $k$ for identification of one cluster in the symmetric $\kNN$ graph)}
  \label{cor:optimal-k-noise-symmetric}
  We use the same notation as in Theorem \ref{cor:optimal-k-noise} and
  define $\maxminprobu = \min_{i=1,\ldots,m} \minprobu\supi$.
  Then all statements about the optimal rates for $k$ in
  Theorem \ref{cor:optimal-k-noise} can be carried over to the symmetric
  $\kNN$ graph, provided one replaces $\minprobu\supi$ with $\maxminprobu$
  in the definitions of $\Gamma\supi$,
  $\Omega\supi_{\textrm{noisefree}}$
  and $\Omega\supi_{\textrm{noisy}}$.
  If Condition~1 holds and the condition 
  $k \leq (n-1) \maxminprobu /2 - 2 \log (n)$
  replaces the corresponding one in Condition~1, we have in the noise-free case
  for sufficiently large $n$
  \begin{displaymath}
    \Pr \big(\textrm{$C\supi$ is identified in $\Gsym(n,k)$}\big)
    \geq 1 - (m+2) e^{-(n-1)\Omega\supi_{\textrm{noisefree}}}.
  \end{displaymath}
  If additionally Condition~2 holds we have in the noisy case
  for sufficiently large $n$
  \begin{displaymath}
    \Pr \big( \textrm{$C\supi$ roughly identified in
      $\tilde{G}\sym \left( n, k, t - \eps, \delta \right)$} \big)
    \geq 1 - (m+7) e^{-(n-1)\Omega\supi_{\textrm{noisy}}}.
  \end{displaymath}
\end{theorem}

Observe that the constant $\minprobu\supi$ has now been replaced by
the minimal $\minprobu^{(j)}$ among all clusters $C\supj$.
This means that the rate
of convergence for the symmetric $\kNN$ graph is governed by the
constant  $\minprobu^{(j)}$ of the ``worst'' cluster,
that is the one which is most difficult to identify. Intuitively, this
worst cluster is the one which has the smallest distance to its
neighboring clusters. In contrast to the results for the mutual
$\kNN$ graph, the rate for identification of $C\supi$ in the symmetric graph is governed by the worst
cluster instead of the cluster $C\supi$ itself.
This is a big disadvantage if the goal is to only identify the
``most significant'' clusters. For this purpose the mutual graph
has a clear advantage.

On the other hand as we will see in the next theorem that the difference
in behavior between the mutual and symmetric graph vanishes as soon as we attempt
to identify {\em all} clusters.

\begin{theorem} {\bf (Optimal $k$ for identification of all clusters in the mutual $\kNN$ graph)}
  \label{thm:all-cluster-identification}
  We use the same notation as in Theorem \ref{cor:optimal-k-noise} and define $\maxminprobu = \min_{i=1,\ldots,m} \minprobu\supi$,
  $p_{\max}=\max_{i=1,\ldots,m} p\supi_{\max}$.
  The optimal choice of $k$ for the identification of all clusters in the mutual $\kNN$ graph
  in $\Gmut(n,k)$ (noise-free case) resp. rough identification of all clusters in
  $\tilde{G}\mut \left( n, k, t - \eps, \delta \right)$ (noisy case) is given by
  \[ k=(n-1)\Gamma^{\textrm{all}} + 1, \quad \textrm{with} \quad  \Gamma^{\textrm{all}}= \frac{\maxminprobu}{2 + \frac{1}{4^d}\frac{t}{p_{\max}}},\]
  provided this choice of $k$ fulfills Condition~1 for all clusters $C^{(i)}$.
  In the noise-free case we get the rate
  \[ \Omega_{\textrm{noisefree}} =\frac{\maxminprobu}{2\,4^{d+1}\,\frac{p_{\max}}{t} + 4},\]
  such that for sufficiently large $n$
  \[ \Pr \big( \textrm{All clusters exactly identified in $\Gmut(n,k)$} \big) \geq 1 - 3m\,e^{-(n-1)\Omega_{\textrm{noisefree}}}.\]
  For the noisy case, assume that additionally Condition~2 holds for all clusters 
  and let $\hat{p}_n$ be a kernel density estimator with bandwidth $h$.
  Then there exist constants $C_1,C_2$ such that if $h^2\leq C_1\eps$ we get with
  \[ \Omega_{\textrm{noisy}} = \min\bigg\{\frac{\maxminprobu}{2\,4^{d+1}\,\frac{p_{\max}}{t} + 4}, \; \frac{n}{n-1}\frac{\delta}{8}, \; \frac{n}{n-1}C_2\, h^d\,\eps^2\bigg\}\]
  and for sufficiently large $n$
  \[ \Pr \big( \textrm{All clusters roughly ident. in $\tilde{G}\mut \left( n, k, t - \eps, \delta \right)$} \big) \geq 1 - (3m+5)\,e^{-(n-1)\Omega_{\textrm{noisy}}}.\]
\end{theorem}

We can see that as in the previous theorem, the constant which now
governs the speed of convergence is the worst case constant among all
the $\minprobu\supj$. In the setting where we want to identify all
clusters this is unavoidable. Of course the identification of
``insignificant'' clusters will be difficult, and the overall behavior
will be determined by the most difficult case. This is what is
reflected in the above theorem. The corresponding theorem for
identification of all clusters in the symmetric $\kNN$ graph looks
very similar, and we omit it.

So far for the noisy case we mainly considered the case of rough
cluster identification. As we have seen, in this setting the results
of the noise-free case are very similar to the ones in the noisy
case. Now we would like to conclude with a theorem for exact cluster
identification in the noisy case.

\begin{theorem}[Exact identification of clusters in the noisy case] \label{cor:optimal-k-noise-exact}
\sloppy Let $p$ be three times continuously differentiable with uniformly bounded
derivatives and let
$\hat{p}_n$ be a kernel density estimator with bandwidth
$h_n=h_0 (\log n / n)^{1/(d+4)}$ for some $h_0>0$.
For a suitable constant $\eps_0>0$ set
$\eps_n=\eps_0 (\log n / n)^{2/(d+4)}$.
Then there exist constants $c_1,c_2$ such that for $n \to \infty$ and
$c_1\log n \leq k\leq c_2 n$ we obtain
\[ \textrm{Cluster $C\supi$ is exactly identified in $\tilde{G}\mut \left( n, k, t - \eps_n, \delta \right)$ almost surely.}\]
\end{theorem}

Note that as opposed to rough cluster identification, which is a
statement about a given finite nearest neighbor graph, exact cluster
identification is an inherently asymptotic property. The complication
in this asymptotic setting is that one has to balance the speed of
convergence of the density estimator with the one of the ``convergence
of the graph''. The exact form of the density estimation is not important. Every other
density estimator with the same convergence rate would yield the same result. One can
even lower the assumptions on the density to $p \in C^1(\R^d)$ (note that differentiability
is elsewhere required). Finally, note that since it is technically difficult to grasp the graph
after the small components have been discarded, we could not prove what the optimal $k$ in
this setting should be.

\section{Proofs}
\begin{figure}[h!]
  \begin{center}
    \includegraphics[width=0.9\textwidth]{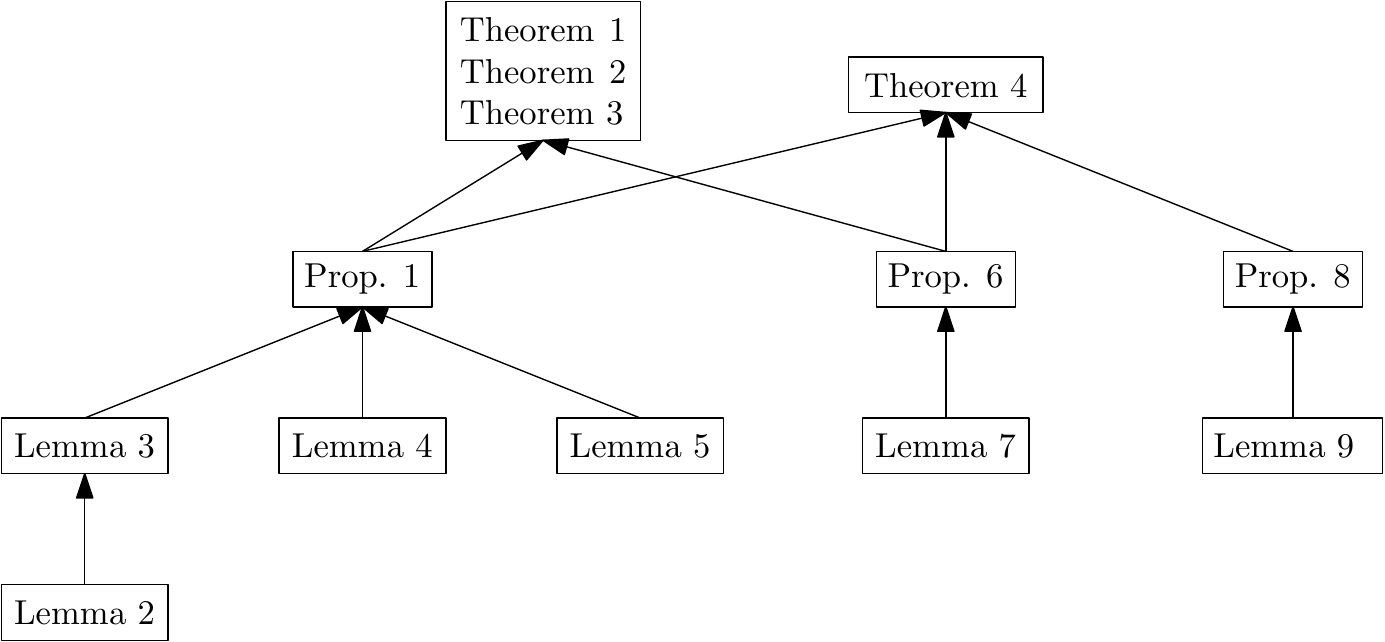}
    \caption{\label{fig:proof_structure} The structure of our proofs.
      Proposition~\ref{pro:super_connectedness} deals with within-cluster
      connectedness and Proposition~\ref{pro:inter-mutual} with
      between-cluster disconnectedness.
      Proposition~\ref{prop:RatioClusterPoints} bounds the ratio of background
      and cluster points for the asymptotic analysis of exact cluster
      identification.}
  \end{center}
\end{figure}

The propositions and lemmas containing the major proof steps
are presented in Section~\ref{sec:main_props}.
The proofs of the theorems themselves can be found in
Section \ref{sec:proof_thms}.
An overview of the proof structure can be seen in
Figure~\ref{fig:proof_structure}.

\subsection{Main propositions for cluster identification} \label{sec:main_props}

In Proposition~\ref{pro:super_connectedness} we identify some
events whose combination guarantee the connectedness of a
cluster in the graph and at the same time that there is not a
connected component of the graph that consists of background points
only. The probabilities of the events appearing in the proposition
are then bounded in Lemma~\ref{lem:conn_within_noise}-\ref{lem:boundary_size}.
In Proposition~\ref{pro:inter-mutual} and Lemma~\ref{pro:asymp_iso}
we examine the probability of connections between clusters.
The section concludes with Proposition~\ref{prop:RatioClusterPoints}
and Lemma~\ref{pro:boundarymass}, which are used in the exact cluster identification
in Theorem~\ref{cor:optimal-k-noise-exact}, and some remarks about
the differences between the noise-free and the noisy case.

\begin{proposition}[Connectedness of one cluster $C\supi$ in the noisy case]
\label{pro:super_connectedness}
Let $\ConnectivityEvent\supi_n$ denote the event that in $\tilde{G}\mut
\left( n, k, t - \eps_n, \delta \right)$ (resp. $\tilde{G}\sym
\left( n, k, t - \eps_n, \delta \right)$) it holds that
\vspace{-0.4cm}
\begin{itemize}
\item
  all the sample points from $C\supi$ are contained in the
  graph,
\item
  the sample points from $C\supi$ are connected
  in the graph,
\item
  there exists no component of the graph which
  consists only of sample points from outside $L(t)$.
\end{itemize}
Then under the conditions
\begin{enumerate}
\item
  $\beta\subi > 2\delta$,
\item
  $\eps_n$ sufficiently small such that
  $\mu \big(\bigcup_i ( C_-\supi(2\eps_n)\backslash C\supi )\big )\leq \delta/2$,
\item
$k\,\geq\,  \kLowerBound$, \\
$k\,\leq\,  \kUpperBoundN \kUpperBoundWithinConnConst$ ,
\end{enumerate}
and for sufficiently large $n$, we obtain
\begin{align*}
  \Pr \big( (\ConnectivityEvent\supi_n)^c \big)
  &\leq \Pr \big( ( \WithinConnEvent_n\supi)^c \big)
  + \Pr \big( ( \ClusterSizeEvent_n\supi )^c \big)
  + \Pr ( \BoundarySizeEvent_n^c)
  + \Pr ( \DensDevEvent_{n}^c ) \\
  &\leq 2 \,e^{-\frac{k-1}{4^{d+1}}\frac{t}{p\supi_{\max}}} + 2
  e^{-n\frac{\delta}{8}} + 2\Pr ( \DensDevEvent_n^c ),
\end{align*}
where the events are defined as follows: \vspace{-0.4cm}
\begin{itemize}
\item
  \sloppy $\WithinConnEvent_n\supi$: the subgraph
  consisting of points from $C^{(i)}$ is connected in
  $G\mut' (n, k, t-\eps_n)$ (resp. $G\sym' (n, k, t-\eps_n)$),
  \item
    $\ClusterSizeEvent_n\supi$: there are more than $\delta n$ sample points from cluster $C^{(i)}$,
  \item
    $\BoundarySizeEvent_n$: there are less than $\delta n$ sample points in the set
    $\bigcup_i \big(C_-\supi(2\eps_n)\backslash C\supi\big)$, and
  \item $\DensDevEvent_n$: $|\hat{p}_n(X_i)-p(X_i)|\leq \eps_n$
    for all sample points $X_i$, $i=1,\ldots,n$.
\end{itemize}
\end{proposition}
\begin{proof}{}
  We bound the probability of $\ConnectivityEvent\supi_n$ using the
  observation that
  $ \WithinConnEvent_n\supi \cap \ClusterSizeEvent_n\supi \cap \BoundarySizeEvent_n
  \cap \DensDevEvent_{n} \subseteq \ConnectivityEvent\supi_n$
  implies
  \begin{align}
    \label{eq:bound_conn_event}
    \Pr \big( (\ConnectivityEvent\supi_n)^c \big)
    &\leq \Pr \big( ( \WithinConnEvent_n\supi)^c \big)
    + \Pr \big( ( \ClusterSizeEvent_n\supi )^c \big)
    + \Pr ( \BoundarySizeEvent_n^c)
    + \Pr ( \DensDevEvent_{n}^c ) .
  \end{align}
  This follows from the following chain of observations. If the
  event $\DensDevEvent_n$ holds, no point with $p(X_i)\geq t$ is
  removed, since on this event $p(X_i) - \hat{p}_n(X_i) \leq
  \eps_n$ and thus $\hat{p}_n(X_i) \geq p(X_i) - \eps_n \geq t
  - \eps_n$,
  which is the threshold in the graph $G'( n, k, t -\eps_n)$.

  If the samples in cluster $C^{(i)}$ are connected in $G'( n, k,
  t-\eps_n )$ ($\WithinConnEvent_n\supi$), and there are more than $\delta n$ samples
  in cluster $C\supi$ ($\ClusterSizeEvent_n\supi$),
  then the resulting component of the graph $G'( n, k, t -\eps_n)$
  is not removed in the algorithm and is thus contained in
  $\tilde{G}( n, k, t - \eps_n, \delta)$.

  Conditional on $\DensDevEvent_n$ all remaining samples are
  contained in $\bigcup_i C_-\supi(2\eps_n)$. Thus all non cluster
  samples lie in $\bigcup_i ( C_-\supi(2\eps_n)\backslash
  C\supi)$. Given that this set contains less than $\delta n$
  samples, there can exist no connected component only consisting of
  non cluster points, which implies that all remaining non cluster points are
  connected to one of the clusters.

  The probabilities for the complements of the events
  $\WithinConnEvent_n\supi$, $\ClusterSizeEvent_n\supi$ and
  $\BoundarySizeEvent_n$ are bounded in Lemmas~3 to~5 below.
  Plugging in those bounds into Equation~\eqref{eq:bound_conn_event}
  leads to the desired result.
\end{proof}
We make frequent use of the following tail bounds for the binomial
distribution introduced by Hoeffding.
\begin{theorem}[Hoeffding, \cite{Hoeffding63}]\label{th:Hoeff}
  Let $M \sim \Bin(n,p)$ and define $\alpha=k/n$. Then,
  \begin{align*}
    \alpha \geq p, \quad \quad &\Pr(M \geq k) \leq e^{-n \, K(\alpha||p)},\\
    \alpha \leq p, \quad \quad &\Pr(M \leq k) \leq e^{-n \, K(\alpha||p)},
  \end{align*}
  where $K(\alpha||p)$ is the Kullback-Leibler divergence of $(\alpha,1-\alpha)$ and $(p,1-p)$,
  \[ K(\alpha||p) = \alpha \log\Big(\frac{\alpha}{p}\Big) + (1-\alpha) \log\Big(\frac{1-\alpha}{1-p}\Big).\]
\end{theorem}
In the following lemmas we derive bounds for the probabilities of
the events introduced in the proposition above.

\begin{lemma}[Within-cluster connectedness $( \WithinConnEvent_n\supi )$]
  \label{lem:conn_within_noise}
  As in Proposition~\ref{pro:super_connectedness} let
  $\WithinConnEvent_n\supi$ denote the event that the points of cluster $C^{(i)}$
  are connected in \sloppy $G\mut' (n, k, \eps_n)$ (resp. $G\sym' (n, k, \eps_n)$).
  For $z \in ( 0, 4 \, \min\{u\supi,\nu^{(i)}_{\max}\} )$,
  \begin{align*} \label{eq:within-connectedness-noise}
    \Pr \big( ( \WithinConnEvent_n\supi )^c \big)
    &\leq n\,\beta\subi\,
    \Pr (M \geq k) +
    N \,\Big( 1 - t \,\eta_d \frac{z^d}{4^d}\Big)^n
    + \Pr \big( \DensDevEvent_{n}^c \big),
  \end{align*}
  where $M$ is a $\Bin (n-1,p^{(i)}_{\max} \eta_d z^d)$-distributed random variable
  and  $N \leq ( 8^d \vol ( C^{(i)} ) )/ ( z^d \eta_d )$.
\end{lemma}

\begin{proof}{}
Given that $\DensDevEvent_n$ holds, all samples lying in cluster $C^{(i)}$
are contained in the graph $G' ( n, k, \eps_n )$. Suppose that we have a covering
of $C^{(i)}\backslash \collarSet^{(i)}(z/4)$ with balls of radius $\frac{z}{4}$.
By construction every ball of the covering lies entirely in $C^{(i)}$, so that
$t$ is a lower bound for the minimal density in each ball.
If every ball of the covering contains at least one sample point and the minimal $\kNN$ radius of samples in $C^{(i)}$ is larger
or equal to $z$, then all samples of $C^{(i)}\backslash \collarSet^{(i)}(z/4)$
are connected in $G'( n, k, \eps_n )$ given that $z \leq 4\nu\supi_{\max}$.
Moreover, one can easily check that all samples lying in the collar set $\collarSet^{(i)}(z/4)$
are connected to $C^{(i)}\backslash \collarSet^{(i)}(z/4)$. In total,
then all samples points lying in $C\supi$ are connected.
Denote by $\F\supi_z$ the event that one ball in the covering with balls of radius $z/4$ contains no sample point.
Formally, $\{ R^{(i)}_{\min} > z \} \cap (\F^{(i)}_z)^c$ implies
connectedness of the samples lying in $C\supi$ in the graph $G'( n, k, \eps_n )$.

Define $N_s = | \{ j \neq s \condon X_j \in B ( X_s, z ) \} |$ for $1 \leq s \leq n$.
Then
$\{R\supi_{\min} \leq z\} = \cup_{s=1}^n \big\{ \{N_s \geq k\} \cap \{X_s \in C\supi\} \big\}$. We have
\begin{align*}
  \Pr\big(R^{(i)}_{\min} \leq z\big)
  &\leq \sum_{s=1}^n \Pr\big(N_s\geq k\,|\, X_s \in C\supi\big)\,\Pr\big(X_s \in C\supi\big)
  \leq n \beta\subi \Pr(U\geq k),
\end{align*}
where $U \sim \Bin ( n-1,\sup_{x \in C\supi}\mu ( B(x,z) )$.
The final result is obtained using the upper bound
$\sup_{x\in C\supi} \mu (B(x,z) )\leq p\supi_{\max} \eta_d z^d$.

For the covering a standard construction using a $z/4$-packing provides us with the covering.
Since $z/4 \leq \nu\supi_{\max}$ we know that balls of
  radius $z/8$ around the packing centers are subsets of $C^{(i)}$ and disjoint by construction.
  Thus, the total volume of the $N$ balls is bounded by the volume of $C\supi$
  and we get $N ( z/8 )^d \eta_d \leq \vol (C^{(i)} )$. Since we assume that
  $\DensDevEvent_n$ holds, no sample lying in $C\supi$ has been discarded. Thus
  the probability for one ball of
  the covering being empty can be upper bounded by $(1-t \,\eta_d \, z^d / 4^d)^n$, where we have used
  that the balls of the covering are entirely contained in $C\supi$ and thus the density is lower
  bounded by $t$. In total, a union bound over all balls in the covering yields,
  \begin{displaymath}
    \Pr\big( \F^{(i)}_z \big) \,\leq\,
    N \,(1-t \,\eta_d \, z^d / 4^d)^n + \Pr \big( \DensDevEvent_{n}^c \big) .
  \end{displaymath}
  Plugging both results together yields the final result.
\end{proof}
In Lemma~\ref{lem:conn_within_noise} we provided a bound on the
probability which includes two competing terms for the choice of $z$. One favors
small $z$ whereas the other favors large $z$. The next lemma will provide a
trade-off optimal choice of the radius $z$ in terms of $k$.
\begin{lemma}[Choice of $k$ for within-cluster connectedness
    $( \WithinConnEvent_n\supi )$] \label{pro:intra-mutual}
  If $k$ fulfills Condition~(3) of Proposition~\ref{pro:super_connectedness},
  we have for sufficiently large $n$
  \begin{displaymath}
    \Pr \big( ( \WithinConnEvent_n\supi )^c \big)
    \leq 2 \,e^{-\frac{k-1}{4^{d+1}}\frac{t}{p\supi_{\max}}}
    + \Pr \left( \DensDevEvent_n^c \right).
  \end{displaymath}
\end{lemma}

\begin{proof}{} The upper bound on the probability of $(\WithinConnEvent_n\supi )^c$ given in Lemma \ref{lem:conn_within_noise} has two
terms dependent on $z$. The tail bound for the binomial distribution is small if $z$ is chosen
to be small, whereas the term from the covering is small given that $z$ is large.
Here, we find a choice for $z$ which is close to optimal.
Define $p=p\supi_{\max}\eta_d z^d$ and $\alpha=k/(n-1)$.
Using Theorem \ref{th:Hoeff} we obtain for $M \sim \Bin(n-1,p)$ and
a choice of $z$ such that $p<\alpha$,
\begin{align*}
  n \beta\subi \Pr(M\geq k) &\leq n \beta\subi e^{-(n-1)\big(\alpha \log\big(\frac{\alpha}{p}\big) + (1-\alpha)\log\big(\frac{1-\alpha}{1-p}\big)\big)}\\
  &\leq n \beta\subi e^{-(n-1)\big(\alpha \log\big(\frac{\alpha}{p}\big)
    + p-\alpha\big)},
\end{align*}
where we have used $\log(z)\geq (z-1)/z$ for $z>0$.
Now, introduce $\theta :=\eta_d z^d/\alpha $ so that $p=p\supi_{\max}\,\theta\, \alpha$,
where with $p\leq \alpha$ we get, $0\leq \theta \,p\supi_{\max}\leq 1$.
Then,
\begin{align}\label{eq:bound1}
  n \beta\subi \Pr(M\geq k)
  & \leq n \beta\subi\,e^{- k \big( \log\big(\frac{1}{p\supi_{\max}\theta}\big)
    + \theta p\supi_{\max} -1\big)}\nonumber\\
  &\leq e^{-\frac{k}{2} \big( \log\big(\frac{1}{p\supi_{\max}\theta}\big)
    + \theta p\supi_{\max} -1\big)},
\end{align}
where we used in the last step an upper bound on the term $n\beta\subi$
which holds given
$k\geq \big(2 \log(\beta\subi n)\big)/\big(\log(1/(\theta p\supi_{\max})) + \theta p\supi_{\max}-1\big)$.
On the other hand,
\begin{align*}
  N(1-t \,\eta_d \,z^d / 4^d)^n
  = N e^{n \log(1-t \,\eta_d \,z^d / 4^d)} \leq N e^{-n \,t \,\eta_d \,z^d / 4^d}
\end{align*}
where we used $\log(1-x)\geq -x$ for $x\leq 1$.
With $\eta_d z^d = \theta \alpha$ and the upper bound on $N$ we get
using $n / (n-1)\geq 1$, %
\begin{align}\label{eq:bound2}
N e^{-n \,t \,\eta_d \,z^d / 4^d} &\leq
e^{-\frac{n \,t \,\theta \,\alpha}{ 4^d}
+ \log\big(\frac{\vol(C\supi)8^d}{\theta\,\alpha}\big)}
\leq  e^{- k \frac{t \,\theta}{4^d}
+ \log\big(\frac{\vol(C\supi)8^d}{\theta\,\alpha}\big)} \nonumber\\
&\leq e^{- k\frac{t \,\theta}{2 \, 4^d}},
\end{align}
where the last step holds given $k \geq \frac{2 \, 4^{d}}{t \theta} \log\Big(\frac{n \vol(C\supi) 8^d }{\theta}\Big)$.
Upper bounding the bound in \eqref{eq:bound1} with the one in \eqref{eq:bound2} requires,
\begin{align*}
  \frac{t\theta}{2\, 4^{d}}
  \leq \frac{1}{2}\big(\log\big(\frac{1}{p\supi_{\max}\theta}\big) + \theta p\supi_{\max} -1\big).
\end{align*}
Introduce, $\gamma=\theta \,p\supi_{\max}$, then this is equivalent to
$\gamma t/(4^{d} p\supi_{\max})  \leq (-\log (\gamma) +\gamma -1 )$.
Note, that $t/(4^{d} p\supi_{\max}) \leq 1/4$.
Thus, the above inequality holds for all $d\geq 1$ given that
$-\log(\gamma) \geq 1-3\gamma/4$.
A simple choice is $\gamma=1/2$ and thus $\theta=1/(2p\supi_{\max})$, which
fulfills $\theta p\supi_{\max}\leq 1$.
In total, we obtain with the result from Lemma~\ref{lem:conn_within_noise},
\begin{displaymath}
  \Pr \big( ( \WithinConnEvent_n\supi )^c \big)
  \leq 2 \,e^{-\frac{k}{4^{d+1}}\frac{t}{p\supi_{\max}}}
  + \Pr \left( \DensDevEvent_n^c \right)
  \leq 2 \,e^{-\frac{k-1}{4^{d+1}}\frac{t}{p\supi_{\max}}}
  + \Pr \left( \DensDevEvent_n^c \right).
\end{displaymath}
We plug in the choice of $\theta$ into the lower bounds on $k$.
One can easily find an upper bound for the maximum of the two lower bounds which gives,
\begin{displaymath}
  k \geq \kLowerBound .
\end{displaymath}
The upper bound, $z \leq 4 \min\{u\supi, \nu\supi_{\max}\}$,
translates into the following upper bound on $k$,
$k \leq \kUpperBoundN \kUpperBoundWithinConnConst$.
\end{proof}
The result of this lemma means that if we choose $k \geq c_1 + c_2 \log n$
with two constants $c_1,c_2$ that depend on
the geometry of the cluster and the respective density, then the
probability that the cluster is disconnected approaches zero
exponentially in $k$.

Note that due to the constraints on the covering radius, we have
to introduce an upper bound on $k$ which depends linearly on
$n$. However, as the probability of connectedness is monotonically
increasing in $k$, the value of the within-connectedness bound for
this value of $k$ is a lower bound for all larger $k$ as well.
Since the lower bound on $k$ grows with $\log n$ and the upper bound
grows with $n$, there exists a feasible region for $k$ if $n$ is large
enough.
\begin{lemma}[Event $\ClusterSizeEvent_n\supi$] \label{lem:cluster_size}
  As in Proposition~\ref{pro:super_connectedness} let
  $\ClusterSizeEvent_n\supi$ denote the event that
  there are more than $\delta n$ sample points from cluster $C^{(i)}$.
  If $\beta\subi > \delta$ then
  \begin{align*}
    \Pr \left( \left( \ClusterSizeEvent_n\supi \right)^c \right)
    &\leq \exp \Big( - \frac{1}{2} n \beta\subi
\Big( \frac{\beta\subi - \delta} {\beta\subi} \Big) ^2 \Big) .
  \end{align*}
\end{lemma}
\begin{proof}{}
 Let $M^{(i)}$ be the number of samples in cluster $C^{(i)}$. Then,
  \begin{align*}
    \Pr \big( M^{(i)} < \delta n\big)
    \leq \Pr \Big( M^{(i)} < \frac{\delta}{\beta\subi}
    \beta\subi n \Big)
    \leq \exp \Big( - \frac{1}{2} n \beta\subi
    \Big( \frac{\beta\subi - \delta}{\beta\subi} \Big)^2 \Big),
  \end{align*}
  where we used $M^{(i)} \sim \Bin ( n , \beta\subi )$ and a Chernoff bound.
\end{proof}

\begin{lemma}[Event $\BoundarySizeEvent_n$] \label{lem:boundary_size}
  As in Proposition~\ref{pro:super_connectedness} let
  $\BoundarySizeEvent_n$ denote the event that there are less than $\delta n$
  sample points in all the boundary sets
  $C^{(j)}_- ( 2 \eps_n ) \setminus C^{(j)}$ together.
  If $\sum_{j=1}^m \mu \big( C^{(j)}_- ( 2 \eps_n ) \setminus C^{(j)} \big)<\delta / 2$,
  we have
  $\Pr ( \BoundarySizeEvent_n^c ) \leq \exp ( - \delta n/8 )$.
\end{lemma}

\begin{proof}{}
  By assumption, for the probability mass in the boundary strips we have
  \sloppy
  $\sum_{j=1}^m \mu ( C^{(j)}_- ( 2 \eps_n ) \setminus C^{(j)} ) < \delta / 2$.
  Then the probability that there are at least $\delta n$ points in the
  boundary strips
  can be bounded by the probability that a $\Bin (n, \delta / 2 )$-distributed
  random variable $V$ exceeds $\delta n$.
  Using a Chernoff bound we obtain
  $\Pr \left( V > \delta n \right) \leq \exp ( -  \delta n  / 8 )$.
\end{proof}
The proposition and the lemmas above are used in the analysis of
within-cluster connectedness. The following proposition deals with
between-cluster disconnectedness.

We say that a cluster $C\supi$ is {\em isolated} if the
subgraph of $\tilde{G}\mut \left( n, k, t - \eps_n, \delta \right)$
(resp. $\tilde{G}\sym \left( n, k, t - \eps_n, \delta \right)$)
corresponding to cluster $C\supi$ is
not connected to another subgraph corresponding to
any other cluster $C^{(j)}$ with $j\neq i$.
Note, that we assume $\min_{j=1,\ldots,m} \dist (C\supi_-(2\eps_n),C\supj_-(2\eps_n))\geq u\supi$
for all $\eps_n \leq \tilde{\eps}$. The following proposition
bounds the probability for cluster isolation.
This bound involves the probability
that the maximal $k$-nearest-neighbor radius is greater than some
threshold. Therefore in Lemma~\ref{pro:asymp_iso} we derive a bound
for this probability.
Note that our previous paper \cite{OurALTPaper:2007} contained an
error in the result corresponding to Lemma~\ref{pro:asymp_iso},
which changed some constants but did not affect the main results.

\begin{proposition}[Cluster isolation] \label{pro:inter-mutual}
Let $\IsolationEvent\supi_n$ denote the event that the subgraph
of the samples in $C_-\supi(2\eps_n)$ is isolated in
$\tilde{G}\mut \left( n, k, t - \eps_n, \delta \right)$.
Then given that $\eps_n \leq \tilde{\eps}$, $k < \minprobu\supi n/2 - 2\log(\tilde{\beta}\subi n)$,
we obtain
\begin{align*}
    \Pr \big( ( \IsolationEvent\supi_n)^c \big)
    &\,\leq\, \Pr\big(\, \tilde{R}^{(i)}_{\max} \geq u^{(i)} \,\big)
    + \Pr \big( \DensDevEvent_{n}^c \big)
    \,\leq\,  e^{-\frac{n-1}{2}\big(\frac{\minprobu^{(i)}}{2}-\frac{k-1}{n-1}\big)}
    + \Pr \big( \DensDevEvent_{n}^c \big) .
\end{align*}
Let $\hat{\IsolationEvent}\supi_n$ be the event that the subgraph of
samples in $C\supi_-(2\eps_n)$ is isolated in
$\tilde{G}\sym(n,k,t-\eps_n,\delta)$.
Define $\maxminprobu=\min\nolimits_{i=1,\ldots,m}\minprobu^{(i)}$ and $\tilde{\beta}_{\max}=\max\nolimits_{i=1,\ldots,m}\tilde{\beta}\subi$.
Then for $\eps_n \leq \tilde{\eps_n}$, $k < \maxminprobu n/2 - 2\log(\tilde{\beta}_{\max}\,n)$,
we obtain
\begin{align*}
  \Pr \big( ( \hat{\IsolationEvent}\supi_n )^c \big)
  &\leq \sum_{j=1}^m \Pr\big(\, \tilde{R}^{(j)}_{\max} \geq u^{(j)} \,\big)
  + \Pr \big( \DensDevEvent_{n}^c \big)
  \leq m  \,e^{-\frac{n-1}{2}\big(\frac{\maxminprobu}{2}-\frac{k-1}{n-1}\big)}
  + \Pr \big( \DensDevEvent_{n}^c \big) .
\end{align*}
\end{proposition}
\begin{proof}{}
We have
$\Pr ( ( \IsolationEvent\supi_n )^c )
\leq \Pr ( ( \IsolationEvent\supi_n )^c \condon \DensDevEvent_{n} )
+ \Pr ( \DensDevEvent_{n}^c )$.
Given the event $\DensDevEvent_{n}$, the remaining points in
$\tilde{G}\mut (n,k,t-\eps_n,\delta)$ are samples from
$C^{(j)}_- (2 \eps_n)$ ($j=1,\ldots ,m$). By assumption we
have for $\eps_n \leq \tilde{\eps}$ that
\sloppy $\min_{j\neq i}\dist (C\supi_-(2\eps_n),C\supj_-(2\eps_n))\geq u\supi$.
In order to have edges from samples in $C\supi_-(2\eps_n)$ to any other part in
$\tilde{G}\mut (n,k,t-\eps_n,\delta)$,
it is necessary that  $\tilde{R}\supi_{\max}\geq u\supi$.
Using Lemma \ref{pro:asymp_iso} we can lower bound the probability of this event.
For the symmetric $\kNN$ graph there can be additional edges
from samples in $C\supi_-(2\eps_n)$ to other parts in
the graph if samples lying in $C\supi_-(2\eps_n)$ are among the $\kNN$-neighbors
of samples in $C^{(j)}_-(2\eps_n)$, $j\neq i$.
Let $u^{ij}$ be the distance between
$C\supi_- (2\tilde{\eps} )$ and $C^{(j)}_- (2 \tilde{\eps})$. There can be edges from
samples in $C^{(i)}_- ( 2\eps_n )$ to any other part in
$\tilde{G}\sym(n,k,\eps_n,\delta)$ if the following event holds:
$\{\tilde{R}_{\max}^{(i)}\geq u^{(i)}\} \cup \{ \cup_{j\neq i} \{
\tilde{R}_{\max}^{(j)}\geq u^{ij}\}\}$. Using a union bound we obtain,
\begin{align*}
    \Pr \big( \big( \hat{\IsolationEvent}\supi_n \big)^c
    \condon \DensDevEvent_{n} \big)
    \leq \Pr\big(\tilde{R}_{\max}^{(i)}\geq u^{(i)}\big)
    + \sum_{j\neq i} \Pr\big(\tilde{R}_{\max}^{(j)}\geq u^{ij}\big) .
\end{align*}
With  $u^{(j)} \leq u^{ij}$ and Lemma~\ref{pro:asymp_iso} we obtain the result for $\tilde{G}\sym(n,k,\eps_n,\delta)$.
\end{proof}
The following lemma states the upper bound for the probability that
the maximum $k$-nearest neighbor radius $\tilde{R}^{(i)}_{\max}$ of samples in
$C\supi_-(2\eps_n)$ used in the proof of Proposition \ref{pro:inter-mutual}.
\begin{lemma}[Maximal $\kNN$ radius] \label{pro:asymp_iso}
Let $k < \minprobu\supi n/2 - 2 \log(\tilde{\beta}\subi n)$. Then
  \begin{displaymath}
    \Pr\big( \tilde{R}^{(i)}_{\max} \geq u^{(i)} \big) \,\leq \, e^{-\frac{n-1}{2}\big(\frac{\minprobu^{(i)}}{2}-\frac{k-1}{n-1}\big)}.
  \end{displaymath}
\end{lemma}
\begin{proof}{}
Define $N_s = | \{ j \neq s \,|\, X_j \in B(X_s, u^{(i)}) \} |$ for
$1 \leq s \leq n$. Then
$\{\tilde{R}^{(i)}_{\max} \geq u^{(i)} \} = \bigcup_{s=1}^n \{ N_s \leq k-1 \; \cap \; X_s \in C^{(i)}_-(2\tilde{\eps})\}$.
Thus,
\begin{align*}
\Pr\big(\tilde{R}^{(i)}_{\max} \geq u^{(i)}\big) \;\leq\; \sum_{s=1}^n \Pr\big(N_s \leq k-1 \,|\, X_s \in C^{(i)}_-(2\tilde{\eps})\big)\, \Pr\big(X_s \in C^{(i)}_-(2\tilde{\eps})\big) .
\end{align*}
Let $M \sim \Bin(n-1,\minprobu^{(i)})$. Then $\Pr (N_s \leq k-1 \,|\, X_s \in C^{(i)}_-(2\tilde{\eps}) ) \leq \Pr (M \leq k-1)$.
Using the tail bound from Theorem \ref{th:Hoeff} we obtain for
$k-1<\minprobu^{(i)} (n-1)$,
\begin{align*}
  \Pr\big(\tilde{R}^{(i)}_{\max} \geq u^{(i)}\big) \;&\leq n \;\tilde{\beta}_{(i)}\, \Pr( M \leq k-1)\\
  &\leq n  \tilde{\beta}_{(i)}\, e^{-(n-1)\big(\frac{\minprobu^{(i)}}{2}-\frac{k-1}{n-1}\big)}
  \leq e^{-\frac{n-1}{2}\big(\frac{\minprobu^{(i)}}{2}-\frac{k-1}{n-1}\big)}
\end{align*}
where we use that $\log(x)\geq (x-1) /x$, that
$-w/e$ is the minimum of $x\log(x/w)$ attained at $x=w/e$ and $(1-1/e)\geq 1/2$.
Finally, we use that under the stated condition on $k$ we have
$\log(n \tilde{\beta}_{(i)})\leq [(n-1) \minprobu^{(i)} / 2 - (k-1)] / 2$.

\end{proof}

The following proposition quantifies the rate of \emph{exact cluster
identification}, that means how fast the fraction of points from
outside the level set $L(t)$ approaches zero.
\begin{proposition}[Ratio of boundary and cluster points]
  \label{prop:RatioClusterPoints}\sloppy
  Let $N_{\Cluster}$ and $N_{\NoCluster}$ be the number of cluster points
  and background points in
  $\tilde{G}\mut \left( n, k, t - \eps_n, \delta \right)$
  (resp. $\tilde{G}\sym \left( n, k, t - \eps_n, \delta \right)$)
  and let $\ConnectivityEvent_n^{\text{all}}$
  denote the event that the points of each cluster form a connected
  component of the graph.
  Let $\eps_n \to 0$ for $n \to \infty$ and define
  $\beta=\sum_{i=1}^m \beta\subi$. Then there exists a constant
  $\bar{D}>0$ such that for sufficiently large $n$,
  \begin{align*}
    \Pr\Big( N_{\NoCluster} / N_{\Cluster} \, >\, 4\frac{\bar{D}}{\beta}\eps_n
    \condon \ConnectivityEvent_n^{\text{all}} \Big)
    \leq e^{-\frac{1}{4}\bar{D}\eps_n n} + e^{-n \frac{\beta}{8}}
    + \Pr(\DensDevEvent_n^c).
  \end{align*}
\end{proposition}

\begin{proof}{}
  According to Lemma~\ref{pro:boundarymass} we can find constants
  $\bar{D}\supi>0$
  such that $\mu ( C\supi_-(2 \eps_n) \backslash C\supi ) \leq \bar{D}\supi \eps_n$
  for $n$ sufficiently large, and set $\bar{D}=\sum_{i=1}^m \bar{D}\supi$.
  Suppose that $\DensDevEvent_n$ holds. Then the only points which
  do not belong to a cluster lie in the set $\cup_{i=1}^m C\supi_-(2
  \eps_n) \backslash C\supi$. Some of them might be discarded, but
  since we are interested in proving an upper bound on
  $N_\NoCluster$ that does not matter. Then with
  $p={\Exp N_{\NoCluster}}/{n}\leq \bar{D}\eps_n$ and
  $\alpha = 2\bar{D}\eps_n$ we obtain with Theorem~\ref{th:Hoeff}
  and for sufficiently small $\eps_n$,
  \begin{displaymath}
    \Pr(N_{\NoCluster}\geq 2\bar{D}\eps_n n
    \condon \ConnectivityEvent_n^{\text{all}}, \DensDevEvent_n )
    \leq e^{-n K(\alpha||p)} \leq e^{-n\,\eps_n\,\bar{D}\,(2\log(2)-1)} ,
  \end{displaymath}
  where $K$ denotes the Kullback-Leibler divergence.
  Here we used that for $p\leq \bar{D}\eps_n$ we have $K(\alpha||p) \geq
  K(\alpha||\bar{D}\eps_n)$ and with $\log(1+x) \geq x/(1+x)$ for
  $x>-1$ we have $K(2\bar{D}\eps_n||\bar{D}\eps_n)\geq \bar{D}\eps_n (2\log 2-1)\geq
  \bar{D}\eps_n/4$. Given that $\DensDevEvent_n$ holds and the points of
  each cluster are a connected component of the graph, we know that all
  cluster points remain in the graph and we have
  \begin{displaymath}
    \Pr\big(N_{\Cluster}\leq \frac{\beta n}{2} \condon
    \ConnectivityEvent_n^{\text{all}}, \DensDevEvent_n \big)
    \leq e^{-n \frac{\beta}{8}}
  \end{displaymath}
  using Theorem~\ref{th:Hoeff} and similar arguments to above.
\end{proof}

\begin{lemma}
  \label{pro:boundarymass}
  Assume that $p \in C^2(\R^d)$ with $\norm{p}_\infty= p_{\max}$ and that
  for each $x$ in a
  neighborhood of $\{ p = t \}$ the gradient of $p$ at $x$ is non-zero,
  then there exists a constant $\bar{D}\supi>0$ such that for $\eps_n$
  sufficiently small,
  \[\mu \big( C_-^{(i)} (2 \eps_n) \setminus C^{(i)} \big) \leq \bar{D}\supi \eps_n.\]
\end{lemma}

\begin{proof}{}
  Under the conditions on the gradient and $\eps_n$ small enough, one has
  $C_-^{(i)} ( 2 \eps_n ) \subseteq C^{(i)} + C_1 \eps_n B(0,1)$
  for some constant $C_1$. Here $"+"$ denotes set addition,
  that is for sets $A$ and $B$
  we define $A + B = \{a + b \; | \;  a \in A,\, b \in B\}$.
  Since the boundary $\partial C^{(i)}$ is a smooth $(d-1)$-dimensional
  submanifold in $\R^d$ with a minimal curvature radius $\kappa^{(i)}>0$,
  there exists $\gamma_1>0$ and a constant $C_2$ such that
  $\vol ( C^{(i)} + \eps_n B(0,1) ) \leq \vol (C^{(i)})
  + C_2 \eps_n \vol (\partial C\supi)$
  for $\eps_n < \gamma_1$ (see Theorem~3.3.39 in~\cite{Federer:1969}).
  Thus, by the additivity of the volume,
  \begin{align*}
    \vol \big( C_-^{(i)} ( 2 \eps_n ) \setminus C^{(i)} \big)
    &\leq \vol \big( C^{(i)} + C_1 \eps_n B(0,1) \big) - \vol \big( C^{(i)} \big) \\
    &= C_1 C_2 \vol\big(\partial C\supi \big) \eps_n .
  \end{align*}
  Since $p$ is bounded, we obtain,
  $\mu ( C_-^{(i)} ( 2 \eps_n ) \setminus C^{(i)} )
  \leq C_1 \,C_2\, \vol (\partial C\supi) \,p_{\max}\, \eps_n$,
  for $\eps_n$ small enough. Setting
  $\bar{D}\supi = C_1 \,C_2\, \vol(\partial C\supi) \,p_{\max}$
  the result follows.
\end{proof}

{\bf Noise-free case as special case of the noisy one. } In the
noise-free case, by definition all sample points belong to a
cluster. That means \vspace{-0.4cm}
\begin{itemize}
\item we can omit the density estimation step, which was used to
  remove background points from the graph, and drop the event
  $\DensDevEvent_n$ everywhere,
\item we work with $L(t)$ directly instead of $L(t - \eps)$,
\item we do not need to remove the small components of size
smaller
  than $\delta n$, which was needed to get a grip on the ``boundary''
  of $L(t-\eps)\setminus L(t)$ .
\end{itemize}
In particular, setting $\delta=0$ we trivially have
$\Pr ( ( \ClusterSizeEvent_n\supi )^c ) = 0$ and
$\Pr ( \BoundarySizeEvent_n^c ) = 0$ for all
$i=1, \ldots,m$ and all $n \in \N$.

As a consequence, we can directly work on the graphs $\Gmut (n,k)$ and
$\Gsym (n,k)$, respectively. Therefore, the bounds we gave in the
previous sections also hold in the simpler noise-free case and can
be simplified in this setting.

\subsection{Proofs of the main theorems} \label{sec:proof_thms}
\begin{proof}{ of Theorem~\ref{cor:optimal-k-noise}}
  Given we work on the complement of the event $\IsolationEvent\supi_n$ of
  Proposition~\ref{pro:inter-mutual}, there are no connections in
  $\tilde{G}\mut \left( n, k, t - \eps, \delta \right)$
  between the subgraph containing the points of cluster $C\supi$ and
  points from any other cluster. Moreover, by Proposition
  \ref{pro:super_connectedness} we know that the event
  $\ConnectivityEvent\supi_n=\WithinConnEvent_n\supi \cap
  \ClusterSizeEvent_n\supi \cap \BoundarySizeEvent_n \cap
  \DensDevEvent_n$ implies that the subgraph of all the sample
  points lying in cluster $C\supi$ is connected and all other sample
  points lying not in in the cluster $C\supi$ are either discarded
  or are connected to the subgraph containing all cluster points.
  That means we have identified cluster $C\supi$. Collecting the
  bounds from Proposition \ref{pro:inter-mutual} and
  \ref{pro:super_connectedness}, we obtain
  \begin{align*}
    &\Pr\big(\textrm{Cluster $C\supi$ not roughly identified in
      $\tilde{G}\mut \left( n, k, t - \eps, \delta \right)$}\big)\\
    &\qquad \leq \Pr\big( (\IsolationEvent\supi_n)^c \big)
    + \Pr\big((\ConnectivityEvent\supi_n)^c\big)\\
    &\qquad \leq \Pr\big( (\IsolationEvent\supi_n)^c \big)
    + \Pr \big( ( \WithinConnEvent_n\supi)^c \big)
    + \Pr \big( ( \ClusterSizeEvent_n\supi )^c \big)
    + \Pr ( \BoundarySizeEvent_n^c)
    + \Pr ( \DensDevEvent_{n}^c ) \\
    &\qquad \leq e^{-\frac{n-1}{2}\big(\frac{\minprobu^{(i)}}{2}-\frac{k-1}{n-1}\big)}
    + 2 \,e^{-\frac{k-1}{4^{d+1}}\frac{t}{p\supi_{\max}}} + 2 e^{-n\frac{\delta}{8}}
    + 3 \Pr (\DensDevEvent_n^c ).
  \end{align*}
In the noise-free case the events $\ClusterSizeEvent_n\supi$,
$\BoundarySizeEvent_n$ and $\DensDevEvent_n$ can be ignored.
The optimal choice for $k$ follows by equating the exponents
of the bounds for $(\IsolationEvent\supi_n)^c$ and $( \WithinConnEvent_n\supi)^c$
and solving for $k$. One gets for the optimal $k$,
\[ k = (n-1) \frac{\minprobu\supi}{2 + \frac{1}{4^d}\frac{t}{p_{\max}\supi}} + 1,
\text{ and a rate of }
(n-1)\frac{\minprobu\supi}{2\, 4^{d+1}\frac{p_{\max}\supi}{t} + 4} .
\]

In the noisy case, we know that for $n$ sufficiently large we
can take $\eps$ small enough ($\eps$ is small and fixed) such that
the condition %
$\sum_{j=1}^m \mu ( C^{(j)}_- ( 2 \eps ) \setminus C^{(j)} ) < \delta / 2$
holds.
It is well known that under our conditions on $p$ 
there exist constants $C_1, C_2$ such that
$\Pr(\DensDevEvent_n^c) \leq e^{-C_2 nh^d\eps^2}$
given $h^2 \leq C_1\eps$ (cf.~\citet{Rao:1983}). Plugging this
result into the bounds above the rate of convergence is determined by
the worst exponent,
\[ \min\Big\{\frac{(n-1)\minprobu^{(i)}}{4} -\frac{k-1}{2}\;, \frac{k-1}{4^{d+1}}\frac{t}{p\supi_{\max}},\; n\frac{\delta}{8},\; C_2 nh^d\eps^2\Big\}.\]
However, since the other bounds do not depend on $k$ the optimal choice for $k$ remains the same.
\end{proof}

\begin{proof}{ of Theorem~\ref{cor:optimal-k-noise-symmetric}}
Compared to the proof for cluster identification in the mutual $\kNN$ graph
in Theorem \ref{cor:optimal-k-noise} the only part which changes is the
connectivity event. Here we have to replace the bound on
$\Pr ( (\IsolationEvent\supi_n)^c )$ by the bound on
$\Pr ( (\hat{\IsolationEvent}\supi_n)^c )$
from Proposition \ref{pro:inter-mutual}.
With $\maxminprobu=\min_{i=1,\ldots,m} \minprobu\supi$ we obtain
\begin{align*}
  &\Pr \big( ( \hat{\IsolationEvent}\supi_n )^c \big)
  \leq m \,e^{-\frac{n-1}{2}\big(\frac{\maxminprobu}{2}-\frac{k-1}{n-1}\big)}
  + \Pr(\DensDevEvent^c_n).
\end{align*}
Following the same procedure as in the proof of
Theorem~\ref{cor:optimal-k-noise} provides the result
(for both, the noise-free and the noisy case).
\end{proof}

\begin{proof}{ of Theorem~\ref{thm:all-cluster-identification}}
  We set $\ConnectivityEvent_n^{\text{all}}=\bigcap_{i=1}^m
  \ConnectivityEvent\supi_n$ and $\IsolationEvent_n^{\text{all}}
  =\bigcap_{i=1}^m \IsolationEvent\supi_n$.
  By a slight modification of the proof of Proposition~\ref{pro:super_connectedness}
  and $p_{\max}=\max_{i=1,\ldots,m}p\supi_{\max}$
  \begin{align*}
    \Pr \big( (\ConnectivityEvent_n^{\text{all}})^c \big)
    &\leq 2 \,\sum_{i=1}^m\, e^{-\frac{k-1}{4^{d+1}}\frac{t}{p\supi_{\max}}}
    + 2 e^{-n\frac{\delta}{8}} + 2 \Pr (\DensDevEvent_n^c )\\
    &\leq 2 \,m \,e^{-\frac{k-1}{4^{d+1}}\frac{t}{p_{\max}}} + 2 e^{-n\frac{\delta}{8}}
    + 2 \Pr (\DensDevEvent_n^c ).
  \end{align*}
  By a slight modification of the proof of Proposition~\ref{pro:inter-mutual}
  with $\maxminprobu = \min_{i=1,\ldots,m} \minprobu\supi$,
  \begin{align*}
    \Pr \big( (\IsolationEvent_n^{\text{all}})^c \big)
    \,\leq\, \sum_{i=1}^m e^{-\frac{n-1}{2}\big(\frac{\minprobu^{(i)}}{2}-\frac{k-1}{n-1}\big)}
    + \Pr (\DensDevEvent_n^c)
    \,\leq\, m \,e^{-\frac{n-1}{2}\big(\frac{\maxminprobu}{2}-\frac{k-1}{n-1}\big)}
    + \Pr (\DensDevEvent_n^c) .
  \end{align*}
  Combining these results we obtain
  \begin{align*}
    &\Pr\Big(\textrm{Not all Clusters $C\supi$ roughly identified in
      $\tilde{G}\mut \left( n, k, t - \eps, \delta \right)$}\Big)\\
    &\qquad \leq m \,e^{-\frac{n-1}{2}\big(\frac{\maxminprobu}{2}-\frac{k-1}{n-1}\big)}
    + 3 \Pr (\DensDevEvent_n^c ) + 2\, m \,e^{-\frac{k-1}{ 4^{d+1}}\frac{t}{p_{\max}}}
    + 2 e^{-n\frac{\delta}{8}}.
  \end{align*}
  The result follows with a similar argumentation to the proof of
  Theorem~\ref{cor:optimal-k-noise}.
\end{proof}

\begin{proof}{ of Theorem~\ref{cor:optimal-k-noise-exact}}
  Clearly we can choose $\eps_0>0$ such that $h_n^2 \leq C \eps_n$
  for a suitable constant $C>0$. Then there exists a constant
  $C_2>0$ with $\Pr(\DensDevEvent^c_n) \leq e^{-C_2 n h_n^d \eps_n^2}$.
  Since
  \begin{align*}
    n h_n^d \eps_n^2
    &= h_0^d \eps_0^2 n \big( \frac{\log n}{n} \big)^{\frac{d}{d+4}}
    \big( \frac{\log n}{n} \big)^{\frac{4}{d+4}}
    = h_0^d \eps_0^2 \log n
  \end{align*}
  we have $\sum_{n=1}^\infty \Pr(\DensDevEvent^c_n)<\infty$.
  Moreover, let $\ConnectivityEvent_n^{\text{all}}$
  denote the event that the points of each cluster form a connected
  component of the graph. Then it can be easily checked with
  Proposition~\ref{prop:RatioClusterPoints} that
  we have $\sum_{n=1}^\infty \Pr ( N_{\NoCluster} / N_{\Cluster}
  > 4 \bar{D} \eps_n / \beta  \condon
  \ConnectivityEvent_n^{\text{all}} )<\infty$.
  Moreover, similar to the proof of
  Theorem~\ref{thm:all-cluster-identification}
  one can show that there are constants $c_1,c_2 >0$ such that
  for $c_1 \log n \leq k \leq c_2 n$ cluster $C\supi$ will
  be roughly identified almost surely as $n\rightarrow \infty$.
  (Note here that the bounds on $k$ for which our probability bounds
  hold are also logarithmic and linear, respectively, in $n$).
  Thus, the event $\ConnectivityEvent_n^{\text{all}}$ occurs almost surely
  and consequently $N_{\NoCluster}/N_{\Cluster} \rightarrow 0$ almost surely.
\end{proof}

\section{Discussion}

In this paper we studied the problem of cluster identification in
$\kNN$ graphs. As opposed to earlier work
(\citet{BriChaQuiYuk1997}, \citet{BiaCadPel:2007}) which was only
concerned with establishing connectivity results for a certain
choice of $k$ (resp. $\eps$ in case of an $\eps$-neighborhood
graph), our goal was to determine for which value of $k$ the
probability of cluster identification is maximized. Our work goes
considerably beyond \citet{BriChaQuiYuk1997} and
\citet{BiaCadPel:2007}, concerning both the results and the proof
techniques. In the noise-free case we come to the surprising
conclusion that the optimal $k$ is rather of the order of $c \cdot
n$ than of the order of $\log n$ as many people had suspected,
both for mutual and symmetric $\kNN$ graphs. A similar result also
holds for rough cluster identification in the noisy case. Both
results were quite surprising to us --- our first naive
expectation based on the standard random geometric graph
literature had been that $k \sim \log n$ would be optimal. In
hindsight, our results perfectly make sense. The minimal $k$ to
achieve within-cluster connectedness is indeed of the order $\log
n$. However, clusters can be more easily identified the tighter
they are connected. In an extreme case where clusters have a very
large distance to each other, increasing $k$ only increases the
within-cluster connectedness. Only when the cluster is fully
connected (that is, $k$ coincides with the number of points in the
cluster, that is $k$ is a positive fraction of $n$),  connections
to other clusters start to arise. Then the cluster will not be
identified any more. Of course, the standard situation will not be
as extreme as this one, but our proofs show that the tendency is
the same.

While our results on the optimal choice of $k$ are nice in theory,
in practical application they are often hard to realize. The
higher the constant $k$ in the $\kNN$ graph is chosen, the less
sparse the neighborhood graph becomes, and the more resources we
need to compute the $\kNN$ graph and to run algorithms on it. This
means that one has to make a trade-off: even if in many
applications it is impossible to choose $k$ of the order of $c
\cdot n$ for computational restrictions, one should attempt to
choose $k$ as large as one can afford, in order to obtain the most
reliable clustering results.

When comparing the symmetric and the mutual $\kNN$ graph, in terms
of the within-cluster connectedness both graphs behave
similar. But note that this might be an artifact of our proof
techniques, which are very similar in both cases and do not really
make use of the different structure of the graphs. Concerning the
between-cluster disconnectedness, however, both graphs behave very
differently. To ensure disconnectedness of one cluster $C^{(i)}$
from the other clusters in the mutual $\kNN$ graph, it is enough
to make sure that the nearest neighbor of all points of $C^{(i)}$
are again elements of $C^{(i)}$. In this sense, the
between-cluster disconnectedness of an individual cluster in the
mutual graph can be expressed in terms of properties of this
cluster only. In the symmetric $\kNN$ graph this is different.
Here it can happen that some other cluster $C^{(j)}$ links inside
$C^{(i)}$, no matter how nicely connected $C^{(i)}$ is. In
particular, this affects the setting where the goal is to identify
the most significant cluster only. While this is easy in the
mutual $\kNN$ graph, in the symmetric $\kNN$ graph it is not
easier than identifying all clusters as the between-cluster
disconnectedness is governed by the worst case.

From a technical point of view there are some aspects about our
work which could be improved. First, we believe that the geometry
of the clusters does not influence our bounds in a satisfactory
manner. The main geometric quantities which enter our bounds are
simple things like the distance of the clusters to each other, the
minimal and maximal density on the cluster, and so on. However,
intuitively it seems plausible that cluster identification depends
on other quantities as well, such as the shapes of the clusters
and the relation of those shapes to each other. For example, we
would expect cluster identification to be more difficult if the
clusters are in the forms of concentric rings than if they are
rings with different centers aligned next to each other. Currently
we cannot deal with such differences. Secondly, the covering
techniques we use for proving our bounds are not well adapted to
small sample sizes. We first cover all clusters completely by
small balls, and then require that there is at least one sample
point in each of those balls. This leads to the unpleasant side
effect that our results are not valid for very small sample size
$n$. However, we did not find a way to circumvent this
construction. The reason is that as soon as one has to prove
connectedness of a small sample of cluster points, one would have
to explicitly construct a path connecting each two points. While
some techniques from percolation theory might be used for this
purpose in the two-dimensional setting, we did not see any way to
solve this problem in high-dimensional spaces.

In the current paper, we mainly worked with the cluster definition
used in the statistics community, namely the connected components
of $t$-level sets. In practice, most people try to avoid to
perform clustering by first applying density estimation ---
density estimation is inherently difficult on small samples, in
particular in high-dimensional spaces. On the other hand, we have
already explained earlier that this inherent complexity of the
problem also pays off. In the end, not only have we detected where
the clusters are, but we also know where the data only consists of
background noise.

In the computer science community, clustering is often solved via
partitioning algorithms such as mincuts or balanced cuts. Now we
have treated the case of the level sets in this paper, discussing
the graph partitioning case will be the next logical step.
Technically, this is a more advanced setting. The ingredients are
no longer simple yes/no events (such as ``cluster is connected''
or ``clusters are not connected to each other''). Instead, one has
to carefully ``count'' how many edges one has in different areas
of the graph. In future work we hope to prove results on the
optimal choice of $k$ for such a graph partitioning setting.

\bibliography{bib_connectivity}

\begin{thebibliography}{14}
\providecommand{\natexlab}[1]{#1}
\providecommand{\url}[1]{\texttt{#1}}
\expandafter\ifx\csname urlstyle\endcsname\relax
  \providecommand{\doi}[1]{doi: #1}\else
  \providecommand{\doi}{doi: \begingroup \urlstyle{rm}\Url}\fi

\bibitem[Bettstetter(2002)]{Bettstetter02}
C.~Bettstetter.
\newblock On the minimum node degree and connectivity of a wireless multihop
  network.
\newblock In \emph{MobiHoc '02: Proceedings of the 3rd ACM international
  symposium on Mobile ad hoc networking \& computing}, pages 80--91, New York,
  NY, USA, 2002. ACM.

\bibitem[Biau et~al.(2007)Biau, Cadre, and Pelletier]{BiaCadPel:2007}
G.~Biau, B.~Cadre, and B.~Pelletier.
\newblock A graph-based estimator of the number of clusters.
\newblock \emph{ESIAM: Prob. and Stat.}, 11:\penalty0 272--280, 2007.

\bibitem[Bollobas(2001)]{Bollobas01}
B.~Bollobas.
\newblock \emph{Random Graphs}.
\newblock Cambridge University Press, Cambridge, 2001.

\bibitem[Bollobas and Riordan(2006)]{BolRio06}
B.~Bollobas and O.~Riordan.
\newblock \emph{Percolation}.
\newblock Cambridge Universiy Press, Cambridge, 2006.

\bibitem[Brito et~al.(1997)Brito, Chavez, Quiroz, and Yukich]{BriChaQuiYuk1997}
M.~Brito, E.~Chavez, A.~Quiroz, and J.~Yukich.
\newblock Connectivity of the mutual k-nearest-neighbor graph in clustering and
  outlier detection.
\newblock \emph{Stat. Probabil. Lett.}, 35:\penalty0 33--42, 1997.

\bibitem[Devroye and Lugosi(2001)]{DevLug01}
L.~Devroye and G.~Lugosi.
\newblock \emph{{Combinatorial Methods in Density Estimation}}.
\newblock Springer, New York, 2001.

\bibitem[Federer(1969)]{Federer:1969}
H.~Federer.
\newblock \emph{Geometric Measure Theory}, volume 153 of \emph{Die Grundlehren
  der mathematischen Wissenschaften}.
\newblock Springer-Verlag, 1969.

\bibitem[Hartigan(1981)]{Hartigan81}
J.~Hartigan.
\newblock {Consistency of single linkage for high-density clusters}.
\newblock \emph{J. Amer. Statist. Assoc.}, 76:\penalty0 388--394, 1981.

\bibitem[Hoeffding(1963)]{Hoeffding63}
W.~Hoeffding.
\newblock {Probability inequalities for sums of bounded random variables}.
\newblock \emph{J. Amer. Statist. Assoc.}, 58:\penalty0 13--30, 1963.

\bibitem[Kunniyur and Venkatesh(2006)]{KunVen}
S.~S. Kunniyur and S.~S. Venkatesh.
\newblock Threshold functions, node isolation, and emergent lacunae in sensor
  networks.
\newblock \emph{IEEE Trans. Inf. Th.}, 52\penalty0 (12):\penalty0 5352--5372,
  2006.

\bibitem[Maier et~al.(2007)Maier, Hein, and von Luxburg]{OurALTPaper:2007}
M.~Maier, M.~Hein, and U.~von Luxburg.
\newblock Cluster identification in nearest-neighbor graphs.
\newblock In M.Hutter, R.~Servedio, and E.~Takimoto, editors, \emph{Proc. of
  the 18th Conf. on Algorithmic Learning Theory (ALT)}, pages 196--210.
  Springer, Berlin, 2007.

\bibitem[Penrose(2003)]{Penrose03}
M.~Penrose.
\newblock \emph{Random Geometric Graphs}.
\newblock Oxford University Press, Oxford, 2003.

\bibitem[Rao(1983)]{Rao:1983}
B.~L. S.~P. Rao.
\newblock \emph{Nonparametric Functional Estimation}.
\newblock Academic Press, New York, 1983.

\bibitem[Santi and Blough(2003)]{SanBlo03}
P.~Santi and D.~Blough.
\newblock The critical transmitting range for connectivity in sparse wireless
  ad hoc networks.
\newblock \emph{IEEE Trans. Mobile Computing}, 02\penalty0 (1):\penalty0
  25--39, 2003.

\end{thebibliography}

\end{document}